\documentclass[letterpaper,11pt]{article}
\usepackage{a4wide}

\usepackage{bbm}

\usepackage{graphicx} 

\usepackage{amsmath}
\usepackage{lipsum}
\usepackage{subfig}
\usepackage{pifont}
\usepackage{endfloat}
\usepackage{multirow}
\usepackage{array}
\usepackage{hhline}

\usepackage{booktabs} 
\usepackage{array} 
\usepackage{paralist} 
\usepackage{verbatim} 
\usepackage{subfig} 
\usepackage{comment}
\usepackage{pdflscape}
\usepackage{color}

\makeatletter
\newcommand*{\rom}[1]{\expandafter\@slowromancap\romannumeral #1@}
\makeatother

\usepackage{tikz}
\usetikzlibrary{decorations.pathreplacing}
\usepackage{pgfplots}
\pgfmathdeclarefunction{gauss}{2}{%
  \pgfmathparse{1/(#2*sqrt(2*pi))*exp(-((x-#1)^2)/(2*#2^2))}%
}

\usepackage{sectsty}
\usepackage{ulem}
\allsectionsfont{\sffamily\mdseries\upshape} 

\usepackage{amsfonts}
\let\chapter\section
\usepackage{algorithm2e}
\usepackage{multicol}
\usepackage{units}
\usepackage{authblk}
\usepackage{natbib}

\title{Estimating Time-varying Brain Connectivity Networks from Functional MRI Time Series}
\author[1]{Ricardo Pio Monti}
\author[2]{Peter Hellyer}
\author[2]{David Sharp}
\author[2]{Robert Leech}
\author[1]{Christoforos Anagnostopoulos}
\author[1,3]{Giovanni Montana\footnote{Corresponding author: \tt{giovanni.montana@kcl.ac.uk}}}
\affil[1]{Department of Mathematics, Imperial College London, London SW7 2AZ, UK}
\affil[2]{Computational, Cognitive and Clinical Neuroimaging Laboratory, Imperial College London, The Hammersmith Hospital, London W12 0NN, UK}
\affil[3]{Department of Biomedical Engineering, King's College London, St Thomas' Hospital, London SE1 7EH, UK}
\date{} 

\begin{document}

\maketitle

\begin{abstract}

At the forefront of neuroimaging is the understanding of the functional architecture of the human brain. 
 In most applications functional 
networks are assumed to be stationary, resulting in a single network estimated for the entire time course. However recent results suggest that the
connectivity between brain regions is highly non-stationary even at rest. As a result, there is a need for new brain imaging methodologies that 
comprehensively account for the dynamic nature of functional networks. In this work we propose the Smooth Incremental Graphical 
Lasso Estimation (SINGLE) algorithm which estimates dynamic brain networks from fMRI data. We apply the proposed algorithm to functional MRI data 
from 24 healthy patients performing a Choice Reaction Task to demonstrate the dynamic changes in network structure that accompany a simple but 
attentionally demanding cognitive task. Using graph theoretic measures we show that the properties of the 
Right Inferior Frontal Gyrus 
and the Right Inferior Parietal lobe dynamically change with the task. These regions are frequently reported as playing an
important role in cognitive control. Our results suggest that both these regions
play a key role in the attention and executive function during cognitively demanding tasks and may be fundamental in regulating the balance
between other brain regions.

\end{abstract}

\newpage

\section{Introduction}

The discovery of non-invasive brain imaging techniques has greatly boosted interest in cognitive neuroscience. Specifically, the discovery
of functional Magnetic Resonance Imaging (fMRI) has instigated a revolution by providing a non-invasive and readily accessible method by which
to obtain high quality images of the human brain. While traditional fMRI studies focused exclusively on reporting the behaviour of individual
brain regions independently, there has been a recent shift towards understanding the relationships between distinct brain regions, 
referred to as brain connectivity
\citep{Lindquist2008}. The study of brain connectivity has resulted in fundamental insights such as small-world architecture 
\citep{sporns2004organization, bassett2006small} and the presence of hubs \citep{eguiluz2005scale}.

A cornerstone in the understanding brain connectivity is the notion that connectivity can be represented as a graph or network composed
of a set of nodes interconnected by a set of edges.
This allows for connectivity
to be studied using a rich set of graph theoretic tools \citep{Newman2003, graphReview} and has resulted in widespread use of graph theoretic techniques in 
neuroscience \citep{Fair2009, Achard2006}. The first step when looking to study brain connectivity is to define a set of nodes. This can be achieved in many
ways; in the case of fMRI nodes are often defined as spatial regions of interest (ROIs). Alternatively, Independent Component Analysis (ICA) 
can be employed to determine independent components which are subsequently used as nodes \citep{calhoun2009review}.
It follows that each node is associated with its 
own time course of imaging data. This is subsequently used to estimate the connections between nodes, defined as the edge structure of the network. 
In particular, functional connectivity estimates of the edge structure can be obtained by studying the statistical dependencies between each
of the nodes \citep{Strother1995, Lowe1998,Heuvel2010, Friston2011}. The resulting networks, referred to as functional connectivity networks, are the 
primary focus of this work.

Traditionally functional connectivity networks have been estimated by measuring pair-wise linear dependencies between nodes, quantified by Pearson's 
correlation coefficient \citep{Hutchinson2013, graphReview}. This corresponds to estimating the covariance matrix where each entry corresponds to the 
correlation between a distinct pair of nodes. Partial correlations, summarised in the precision or inverse covariance matrix \citep{Whittaker1990}, have 
also been employed 
extensively \citep{Huang2010, Liu2009, Marrelec2006, Sun2004, pandit2013traumatic, hinne2013structurally}. In this case, the correlations between nodes are inferred once the 
effects of all other units have been removed. Partial correlations are typically preferred to Pearson's correlation coefficient as they have been shown 
to be better suited to detecting changes in connectivity structure \citep{Smith2011, marrelec2009}.

Intrinsically linked to the problem of estimating the functional connectivity structure is the issue of estimating the 
true sparsity of the networks in question. There are numerous studies reporting brain networks to be of varying levels of sparsity.
For example, \cite{bullmore} suggest that connectivity networks have evolved to achieve high efficiency of 
information transfer at a low connection cost, resulting in sparse networks. On the other hand, \cite{markov2013cortical}
propose a high-density model where efficiency is achieved 
via the presence of highly heterogeneous edge strengths between nodes.
Here we pose the level of sparsity as a statistical question to be answered by the data.
Due to the presence of noise, it follows that every entry in the estimated precision or covariance
matrices will be non-zero. This results in dense, unparsimonious networks which are potentially dominated by noise. The two most 
common approaches to addressing this problem involve the use of multiple hypothesis tests or regularisation. The former involves testing each edge 
for statistical significance \citep{nichols2003controlling} while the latter involves the formulation of an objective function which contains
an additional regularisation penalty to encourage sparsity. A popular example of such a penalty is the Graphical 
Lasso penalty \citep{glasso}. This penalises the sum 
of the off-diagonal elements in the precision matrix thus balancing a trade-off between sparsity and goodness-of-fit. Furthermore, in many neuroimaging 
studies it is often the case that the number of parameters to estimate exceeds the number of observations. In such scenarios the use 
of regularisation is required for the formulation of a well-posed problem. Moreover, regularisation in the form of the 
Graphical Lasso penalty encourages 
only the presence of edges which are best supported by the data. 

The aforementioned methods are based on the underlying assumption that functional connectivity networks are not changing over time.
However, there is growing evidence that fMRI data is 
non-stationary \citep{Hutchinson2012, Hellyer2013}; this is especially true in task-based fMRI studies \citep{Chang2010}. As a result there is a clear
need to quantify dynamic changes in network structure over time. 
Specifically, there is a need to estimate a network at each observation in order to accurately quantify temporal diversity. 
To date the most commonly used approach to achieve this goal
involves the use of sliding windows \citep{Hutchinson2013}. Here observations lying within
a time window of fixed length are used to calculate the functional connectivity. This window is them shifted,  allowing
for the estimation of dynamic functional networks.
Examples include \cite{Handwerker2012} who use sliding windows
to quantify dynamic trends in correlation and \cite{esposito2003real} who combine sliding windows with ICA.

While sliding windows are a valuable tool for investigating high-level dynamics of functional connectivity networks there are two main issues 
associated with its use. First, the choice of window length can be a difficult parameter to tune. It is advised to set the window length to be large
enough to allow for robust estimation of network statistics without making it too large, which would result in overlooking interesting short-term 
fluctuations \citep{Sakoglu2010}. Second, the use of sliding windows needs to be 
accompanied by an additional mechanism to determine if variations in 
edge structure are significant. This would result in estimated networks where the edge structure is only reported to change when 
substantiated by evidence in the data. We refer to this quality as temporal homogeneity.
One way this can be achieved is via the use of hypothesis tests, 
as in the recently proposed Dynamic Connectivity Regression (DCR)
algorithm \citep{DCR}.

In this work we are concerned with multivariate statistical methods for inferring dynamic functional connectivity networks from fMRI data. We are particularly
interested in two aspects. First, we wish to obtain individual estimates of brain connectivity at each time point as opposed to a
network for the entire time series. This will allow us to fully characterise the dynamic evolution of networks over time and 
provide valuable insight into brain organisation and cognition. Second, we wish to encourage our estimation procedure to 
produce estimates with the two properties discussed previously; sparsity and temporal homogeneity

In order to achieve these goals we propose a new methodology, the Smooth Incremental Graphical Lasso Estimation (SINGLE) algorithm.
SINGLE can be seen as an extension of sliding windows where the two issues mentioned previously --- sparsity and temporal homogeneity --- are addressed. 
First, we propose an objective method for estimating the window length based on cross-validation. We then introduce the SINGLE algorithm which
is capable of accurately estimating dynamic networks. 
The proposed algorithm is able to estimate dynamic networks by minimising
a penalised loss function. This function contains a likelihood term for each observation together with two penalty terms.
Sparsity is achieved via the introduction of a Graphical Lasso penalty while 
temporal homogeneity is achieved by introducing a penalty inspired by the Fused Lasso \citep{FusedLasso} which effectively penalises the difference between
consecutive networks. We study the ability of the SINGLE algorithm to accurately estimate dynamic random networks resembling fMRI data
and benchmark its performance against 
sliding window based algorithms 
and the DCR algorithm. 

We apply the SINGLE algorithm to data collected from 24 healthy subjects whilst performing a Choice Reaction
Time (CRT) task. During the CRT task, subjects were required to make a rapid visually-cued motor decision. Stimulus
presentation was blocked into five on-task periods, 
each preceding a period where subjects were at rest. As a result, we expect there to be an alternating network structure depending on the task. 
This makes the data set particularly suitable for demonstrating the limitations involved with the assumption of stationarity as well as the capabilities 
of the SINGLE algorithm. 

The remainder of this paper is structured as follows: in Section \ref{methods} we introduce and describe the SINGLE framework and optimisation algorithm 
in detail. In Section \ref{simulations} we present the results of our simulation study and 
in Section \ref{crt_app} we then apply the proposed algorithm to fMRI data collected for 24 healthy 
subjects whilst performing a Choice Reaction Time (CRT) task.

\section{Methods}
\label{methods}

We assume we have observed fMRI time series data denoted by $X_1, \ldots, X_T$, where each vector $X_i \in \mathbb{R}^{1 \times p}$ contains 
the BOLD measurements of $p$ nodes at the  $i$th time point. Throughout the remainder we assume that each $X_i$ follows a multivariate Gaussian
distribution, $X_i \sim \mathcal{N}(\mu_i, \Sigma_i)$. Here the mean and covariance are dependent on time index 
in order to accommodate the 
non-stationary nature of fMRI data.

We aim to infer functional connectivity networks over time by 
estimating the corresponding precision (inverse covariance) matrices 
$\{\Theta_i \}=\{ \Theta_1, \ldots, \Theta_T\}$. Here, $\Theta_i$ encodes the partial 
correlation structure at the $i$th observation \citep{Whittaker1990}. It follows that we can encode $\Theta_i$ as a graph or 
network $G_i$ where the presence of an edge implies a non-zero entry in the corresponding precision matrix and can be 
interpreted as a functional relationship
between the two nodes in question. Thus our objective is equivalent to estimating a sequence of time indexed graphs $\{G_i\}=\{G_1, \ldots, G_T\}$ where each 
$G_i$ summarises the functional connectivity structure at the $i$th observation.

We wish for our estimated graphs $\{G_i\}$ to display the following two properties:
\begin{enumerate}[(a)]
 \item \textbf{Sparsity}: The introduction of sparsity is motivation by two reasons; first, the number of parameters to estimate often exceeds 
 the number of observations. In this case the introduction of regularisation is required in order to formulate a well-posed problem. Second, 
 due to the presence of noise, all entries in the estimated precision matrices will be non-zero. This results in dense, unparsimonious networks that 
 are potentially dominated by noise.

 \item \textbf{Temporal homogeneity}: From a biological perspective, it has been reported that functional connectivity networks exhibit changes due to task based demands 
 \citep{esposito2006independent,fornito2012competitive, fransson2006default, sun2007functional}.
 As a result,  
 we expect the network structure to
 remain constant within a neighbourhood of any observation but to vary over a larger time horizon. 
 This is particularly true for task-based fMRI studies where 
 stimulus presentation is blocked. In light of this, we wish to encourage
 estimated graphs with sparse innovations over time. This ensures that a change in the connectivity between two nodes is only reported when is it
 substantiated by evidence in the data.

\end{enumerate}
 
We split the problem of estimating $\{\Theta_i \}$ into two independent tasks. First we look to obtain local estimates of sample
covariance matrices $S_1, \ldots, S_T$. This
is achieved via the use of kernel functions and discussed in detail in section \ref{sec::cov_matrix_est} below.
Assuming such a sequence exists we wish to estimate the corresponding
precision matrices $\{\Theta_i \}$ with the aforementioned properties while ensuring that each $\Theta_i$ adequately describes 
 the corresponding $S_i$. The latter
is quantified by considering the goodness-of-fit:
\begin{equation}
 f(\{\Theta_i\}) = \sum_{i=1}^T  -\mbox{log det }  \Theta_i + \mbox{trace } ( S_i  \Theta_i),
\end{equation}
which is proportional to the negative log-likelihood. While it would
be possible to estimate $\{\Theta_i\}$ by directly minimising $f$, this would not guarantee either of the properties discussed previously. 
In order to enforce sparsity and temporal homogeneity we introduce the following regularisation penalty:
\begin{equation}
 g_{\lambda_1, \lambda_2}(\{\Theta_i \}) = \lambda_1 \sum_{i=1}^T || \Theta_i||_1 + \lambda_2 \sum_{i=2}^T || \Theta_i -  \Theta_{i-1}||_1.
\end{equation}

Sparsity is enforced by the first penalty term which assigns a large cost to matrices with large absolute values, thus effectively shrinking
elements towards zero.
This can be seen as a convex approximation to the combinatorial problem of selecting the number of 
edges. The second penalty term, regularised by $\lambda_2$, encourages temporal homogeneity by penalising the difference between
consecutive networks. This can be seen as an extension of the Fused Lasso penalty \citep{FusedLasso} from the context of linear regression (i.e., enforcing 
similarities across regression coefficients) to penalising the changes in network structure over time. 

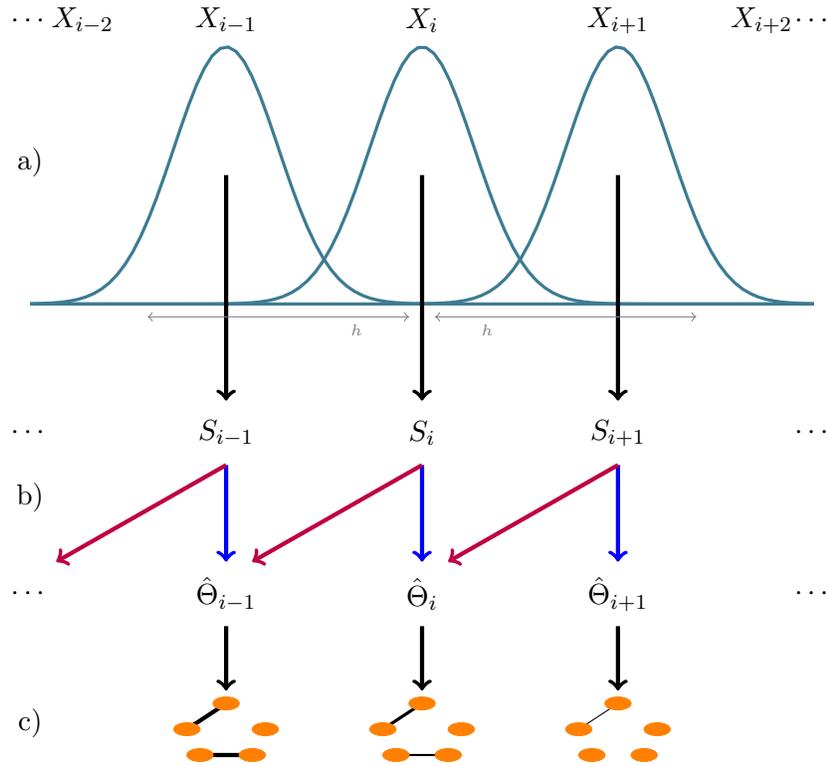
\begin{figure}[ht]
    \centering
\begin{tikzpicture}
\begin{axis}[
  no markers, domain=0:30, samples=100,
  axis lines=none, xlabel=$ $, ylabel=$ $,
  every axis y label/.style={at=(current axis.above origin),anchor=south},
  every axis x label/.style={at=(current axis.right of origin),anchor=west},
  height=5cm, width=12cm,
  xtick=\empty, ytick=\empty,
  enlargelimits=false, clip=false, axis on top,
  grid = major
  ]
  \addplot [very thick,cyan!50!black] {gauss(7.5,2)};
  \addplot [very thick,cyan!50!black] {gauss(15,2)};
  \addplot [very thick,cyan!50!black] {gauss(22.5,2)};
  
  
  \node at (0,220) {$\cdots $};
  \node at (20,220) {$X_{i-2}$};
  \node at (0, 110) {a)};
  \node at (75,220) {$X_{i-1}$};
  \node at (150,220) {$X_{i}$};
  \node at (225,220) {$X_{i+1}$};
  \node at (280,220) {$X_{i+2}$};
  \node at (300,220) {$\cdots$};
  
  \draw [ultra thick, ->] (75, 100) -- (75, -75);
  \draw [ultra thick, ->] (150, 100) -- (150, -75);
  \draw [ultra thick, ->] (225, 100) -- (225, -75);
  
  \draw [gray, <->] ( 45, -10) -- (145, -10);
  \draw [gray, <->] ( 155, -10) -- (255, -10);
  \node [gray] at (125, -20) {\tiny $h$};
  \node [gray] at (175, -20) {\tiny $h$};
  
  \node at (75, -100) {$S_{i-1}$};
  \node at (0, -150) {b)};
  \node at (150, -100) {$S_i$};
  \node at (225, -100) {$S_{i+1}$};
  \node at (0, -100) {$ \cdots $};
  \node at (300, -100) {$ \cdots $};

  \draw [ultra thick, blue, ->] (75, -125) -- (75, -200);
  \draw [ultra thick, blue,->] (150, -125) -- (150, -200);
  \draw [ultra thick, blue,->] (225, -125) -- (225, -200);  
  
  \draw [ultra thick, purple,->] (75, -125) -- (10, -200);
  
  \draw [ultra thick, purple,->] (150, -125) -- (85, -200);
  
  \draw [ultra thick, purple,->] (225, -125) -- (160, -200);
  
  \node at (75, -225) {$\hat \Theta_{i-1}$};
  \node at (150, -225) {$\hat \Theta_i$};
  \node at (0, -325) {c)};
  \node at (225, -225) {$\hat \Theta_{i+1}$};  
  \node at (0, -225) {$ \cdots $};
  \node at (300, -225) {$ \cdots $};

  \draw [ultra thick, ->] (75, -250) -- (75, -300);
  \draw [ultra thick, ->] (150, -250) -- (150, -300);
  \draw [ultra thick, ->] (225, -250) -- (225, -300);


  
  
  \draw [ultra thick] (75, -310) -- (60, -330);
  \draw [ultra thick] (65, -350) -- (85, -350);

  \draw [very thick] (150, -310) -- (135, -330);
  \draw [thick] (140, -350) -- (160, -350);
  
  \draw [thin] (225, -310) -- (210, -330);

  \draw [fill=orange, orange] (75,-310) circle [radius=5];
  \draw [fill=orange, orange] (60,-330) circle [radius=5];
  \draw [fill=orange, orange] (90,-330) circle [radius=5];
  \draw [fill=orange, orange] (65,-350) circle [radius=5];
  \draw [fill=orange, orange] (85,-350) circle [radius=5];

  \draw [fill=orange, orange] (150,-310) circle [radius=5];
  \draw [fill=orange, orange] (135,-330) circle [radius=5];
  \draw [fill=orange, orange] (165,-330) circle [radius=5];
  \draw [fill=orange, orange] (140,-350) circle [radius=5];
  \draw [fill=orange, orange] (160,-350) circle [radius=5];  

  \draw [fill=orange, orange] (225,-310) circle [radius=5];
  \draw [fill=orange, orange] (210,-330) circle [radius=5];
  \draw [fill=orange, orange] (240,-330) circle [radius=5];
  \draw [fill=orange, orange] (215,-350) circle [radius=5];
  \draw [fill=orange, orange] (235,-350) circle [radius=5];  

  
  \end{axis}
  \end{tikzpicture}
  \caption{A graphical representation of the SINGLE algorithm illustrating its various components. a) Gaussian kernels are used to obtain estimate local covariance matrices at each observation. b) These are then used to obtain smooth estimates of precision matrices by combining the Graphical Lasso (blue) and Fused Lasso (purple) penalties. c) Finally the estimated precision matrices can be represented as graphs.  }
    \label{SINGLE_pic}

\end{figure}

The proposed method minimises the following loss function:
\begin{equation}
\label{SINGLE_cost}
 l(\{\Theta_i\}) = f(\{\Theta_i\}) + g_{\lambda_1, \lambda_2}(\{\Theta_i\}).
\end{equation}
This allows for the estimation of time-index precision matrices which display the properties of sparsity and temporal homogeneity while 
providing an accurate representation of the data --- a schematic representation of the proposed algorithm is provided in Figure [\ref{SINGLE_pic}].
The choice of regularisation parameters $\lambda_1$ and $\lambda_2$ allow us to 
balance this trade-off and these are learned from the data as described in section \ref{param_select}. 

The remainder of this section is organised as follows: in section \ref{sec::cov_matrix_est} we describe the estimation of time-varying sample
covariance matrices 
$S_1, \ldots, S_T$ using kernel functions. In section \ref{sec::opt_algo} we outline the optimisation algorithm used to minimise equation (\ref{SINGLE_cost}) as well
as discuss the computational complexity of the resulting algorithm. Finally in section \ref{param_select} we describe how the related parameters can be learnt
from the data and in section \ref{sec::exp_data} we describe the experimental data used in our application.

\subsection{Estimation of Time-varying Covariance Matrices}
\label{sec::cov_matrix_est}

The loss function (\ref{SINGLE_cost}) requires the input of estimated sample covariance matrices $S_1, \ldots, S_T$. 
This is itself a non-trivial and widely studied problem. Under the assumption of stationarity, the covariance matrix can be directly 
calculated as $S = \frac{1}{T-1} \sum_{i=1}^T (X_i - \bar{x})^T (X_i - \bar{x})$ where $\bar{x}$ is the sample mean. However,  when faced with
non-stationary data such an approach is unsatisfactory and there is a need to obtain local estimates of the covariance matrix.

A potential approach involves the use of change-point detection to 
segment the data into piece-wise stationary segments, as is the case in the DCR algorithm \citep{DCR}. Alternatively, a sliding window may be 
used to obtain a locally stationary estimate of the sample covariance at each observation. Due to the sequential nature of the observations,
sliding windows allow us to obtain adaptive estimates by considering only temporally adjacent observations.  

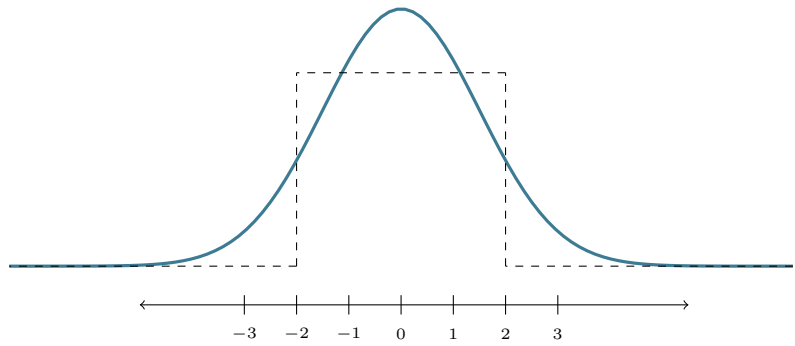
\begin{figure}[ht]
    \centering
\begin{tikzpicture}
\begin{axis}[
  no markers, domain=0:30, samples=100,
  axis lines=none, xlabel=$ $, ylabel=$ $,
  every axis y label/.style={at=(current axis.above origin),anchor=south},
  every axis x label/.style={at=(current axis.right of origin),anchor=west},
  height=5cm, width=12cm,
  xtick=\empty, ytick=\empty,
  enlargelimits=false, clip=false, axis on top,
  grid = major
  ]

  \addplot [very thick,cyan!50!black] {gauss(15,3)};
    
  \draw [dashed] (110, 0) -- (110, 100);
  \draw [dashed] (190, 0) -- (190, 100);
  \draw [dashed] (110, 0) -- (0, 0);
  \draw [dashed] (300, 0) -- (190, 0);
  \draw [dashed] (190, 100) -- (110, 100);
  
  \draw [<->] (50, -20) -- (260, -20);  
  
  \draw (150, -25) -- (150, -15);
  \node at (150, -35) {\tiny $0$};
  
  \draw (110, -25) -- (110, -15);
  \node at (110, -35) {\tiny $-2$};
  
  \draw (190, -25) -- (190, -15);
  \node at (190, -35) {\tiny $2$};
  
  \draw (130, -25) -- (130, -15);
  \node at (130, -35) {\tiny $-1$};
  
  \draw (170, -25) -- (170, -15);
  \node at (170, -35) {\tiny $1$};
  
  \draw (90, -25) -- (90, -15);
  \node at (90, -35) {\tiny $-3$};
  
  \draw (210, -25) -- (210, -15);
  \node at (210, -35) {\tiny $3$};
  
  \end{axis}
  \end{tikzpicture}
  \caption{Example demonstrating the difference between a Gaussian kernel and a sliding window. We note that the sliding window gives zero weighting
  to observations at $\pm 3$ while this is not the case for the Gaussian kernel. We also note that the Gaussian kernel gives greater importance to 
  chronologically adjacent observations while the sliding window gives an equal weighting to all observations within its width.}
    \label{KERNEL_pic}

\end{figure}

A natural extension of sliding windows is to obtain adaptive estimates by downweighting the contribution of past observations. This  
can be achieved using kernel functions. 
Formally, kernel functions have the form $K_h (i,j)$ where $K_h (\cdot, \cdot )$ is a symmetric, 
non-negative function, $h$ is a specified fixed width and $i$ and $j$ are time indices. By considering the uniform kernel,  
$K_h(i,j) = \mathbb{I} \{|i-j|<h \}$\footnote{here $\mathbb{I}(x)$ is the indicator function},
we can see that sliding windows are a special case of kernel functions. This allows us to contrast the behaviour
of sliding windows against alternative kernels, such as the Gaussian kernel:
\begin{equation}
K_h(i,j) = \mbox{exp} \left \{ \frac{-(i-j)^2}{h}\right \}.
\label{gauss_kern}
\end{equation}
Figure [\ref{KERNEL_pic}] provides clear insight into the different behaviour of each of the two kernels. While sliding windows have a sharp cutoff, the Gaussian kernel
gradually reduces the importance given to observations according to their chronological proximity. In this manner, the Gaussian kernel is able to
give greater importance to temporally adjacent observations. In addition to this, sliding windows inherently assume that observations arrive
at equally spaced intervals while the use of more general kernel functions, such as the Gaussian kernel, naturally accommodates
cases where this assumption does
not hold.

Finally, given a kernel function, adaptive estimates of the $i$th sample mean and covariance can be directly calculated as follows:
\begin{align}
 \bar{x}_i &= \frac{ \sum_{j=1}^T K_h(i, j)  X_j}{\sum_{j=1}^T K_h(i, j)}, \label{mean_est}\\
 S_i &= \frac{ \sum_{j=1}^T K_h(i, j)  (X_j-\bar{x}_j)^T (X_j-\bar{x}_j)}{\sum_{j=1}^T K_h(i, j)} \label{cov_est}.
\end{align}

It follows that for both the Gaussian kernel as well as the sliding window the choice of $h$ plays a fundamental role. It is typically advised
to set $h$ to be large enough to ensure robust estimation of covariance matrices without making $h$ too large \citep{Sakoglu2010}. However, data-driven 
approaches are rarely
proposed \citep{Hutchinson2013}. This is partly because the choice of $h$ will depend on many factors, such as the rate of change of the 
underlying networks, which are rarely known apriori.
Here we propose to estimate $h$ using cross-validation. This is discussed in detail in section \ref{param_select}.

\subsection{Optimisation Algorithm}
\label{sec::opt_algo}

Having obtained estimated sample covariance matrices, we turn to the problem of minimising the loss function (\ref{SINGLE_cost}). Whilst this loss 
is convex (see Appendix \ref{app_1}) it is not continuously differentiable due to the presence of the penalty terms. In particular, the presence of the 
Fused Lasso penalty poses a real restriction. Additional difficulty is introduced by the structured nature of the problem: we require that 
each $\Theta_i$ be symmetric and positive definite.

The approach taken here is to exploit the separable nature of equation (\ref{SINGLE_cost}). As discussed previously, the loss function 
is composed of two components; the first of which is proportional to the sum of likelihood terms and the second containing 
the sum of the penalty components. This separability allows us to take advantage of the structure of each component.

There has been a rapid increase in interest in the study of such separable loss functions in the statistics, machine learning and 
optimisation literature. As a result, there are numerous algorithms which can be employed such as Forward-Backward Splitting 
\citep{duchi2009efficient} and Regularised Dual Averaging \citep{xiao2010dual}. 
Here we capitalise on the separability of our problem by implementing an Alternating Directions Method
of Multipliers (ADMM) algorithm \citep{ADMM}. The ADMM is a form 
of augmented Lagrangian algorithm\footnote{see \citet[chap. 4]{bertsekas1999nonlinear} for a concise description of augmented Lagrangian methods} 
which is particularly well suited to dealing with highly structured nature of the
problem proposed here. Moreover, the use of an ADMM algorithm is able to guarantee 
estimated precision matrices, $\{\Theta_i \}$, that are symmetric and positive definite as we outline below.




In order to take advantage of the separability of the loss function (\ref{SINGLE_cost}) we introduce a set of auxiliary variables
denoted $\{Z_i\} = \{Z_1, \ldots, Z_T \}$ where each $Z_i \in \mathbb{R}^{p \times p}$ corresponds to each $\Theta_i$. This allows us to
minimise the loss with respect to each set of variables, $\{\Theta_i\}$ and $\{Z_i\}$ in iterative fashion while enforcing 
an equality constraint on each $\Theta_i$ and $Z_i$ respectively.
Consequently, equation (\ref{SINGLE_cost}) can be reformulated as the 
following constrained minimisation problem:
\begin{align}
\label{sep1}
 \underset{\{\Theta_i\}, \{Z_i\}}{\mbox{minimise}} \hspace{5mm} & \sum_{i=1}^{T} \left ( -\mbox{log det } \Theta_i + \mbox{ trace } ( S_i \Theta_i) \right) + \lambda_1 \sum_{i=1}^{T}||Z_i||_1 + \lambda_2 \sum_{i=2}^T ||Z_i - Z_{i-1}||_1\\
 \label{sep2}
 \mbox{subject to} \hspace{5mm} & \Theta_i = Z_i \hspace{5mm}  i= 1, \ldots, T
\end{align}
where we have replaced $\Theta_i$ with $Z_i$ in both of the penalty terms. 
As a result, $\{\Theta_i\}$ terms are involved only in the likelihood component of equation 
(\ref{sep1}) while $\{Z_i\}$ terms are involved in the penalty components.
This decoupling allows for the  
individual 
structure associated with the $\{\Theta_i\}$ and $\{Z_i\}$ to be leveraged.

The use of an ADMM algorithm requires the formulation of the augmented Lagrangian corresponding to equations (\ref{sep1}) and (\ref{sep2}).
This is defined as:
\begin{align}
\begin{split}
\label{aug_lagrange1}
\mathcal{L}_{\gamma} \left ( \{\Theta_i\}, \{Z_i\}, \{U_i\} \right ) &= -\sum_{i=1}^{T} \left ( \mbox{log det } \Theta_i - \mbox{ trace } (  S_i \Theta_i) \right) + \lambda_1 \sum_{i=1}^{T}||Z_i||_1 \\
&+\lambda_2 \sum_{i=2}^T ||Z_i - Z_{i-1}||_1 +\nicefrac{\gamma}{2} \sum_{i=1}^T \left (|| \Theta_i - Z_i + U_i ||_2^2 - || U_i||_2^2 \right ),
\end{split}
\end{align}
where $\{U_i\} = \{U_1, \ldots, U_T\}$ are scaled Lagrange multipliers such that $U_i \in \mathbb{R}^{p \times p}$.
Equation (\ref{aug_lagrange1}) corresponds to the Lagrangian for equations (\ref{sep1}) and (\ref{sep2}) together with an additional quadratic penalty 
term (see Appendix \ref{app_3} for details).
The latter is multiplied by a constant stepsize parameter $\gamma$ which can typically be set to one. 
The introduction of this term 
is desirable as it often facilitates the minimisation of the Lagrangian; specifically in our case it will make our problem
substantially easier as we outline below.

The proposed estimation procedure works by iteratively minimising equation (\ref{aug_lagrange1}) with respect to each set of variables:  
$\{\Theta_i\}$ and $\{Z_i\}$. This allows us to 
decouple the Lagrangian in such a manner that the individual structure associated with variables $\{\Theta_i\}$ and $\{Z_i\}$ can be 
exploited. 

We write $\{ \Theta^j_i \} = \{\Theta^j_1, \ldots, \Theta^j_T \}$ where $\Theta_i^j$ denotes the estimate of
$\Theta_i$ in the $j$th iteration. The same notation is used for $\{Z_i\}$ and $\{U_i\}$. The algorithm is initialised 
with $\Theta^0_i = I_p$, $Z^0_i = U^0_i = \mbox{\boldmath$0$} \in \mathbb{R}^{p \times p}$ for $i=1, \ldots, T$. At the $j$th iteration of the 
proposed algorithm three steps are performed as follows:

\subsubsection*{Step 1: Update $\{ \Theta^j_i \}$ }

At the $j$th iteration, each $\Theta_i$ is updated independently by minimising equation (\ref{aug_lagrange1}). 
At this step we treat all $\{Z^j_i\}$, $\{U^j_i\}$ and $\Theta_k^j,$ for $ k\neq i$ as constants. As a result, minimising 
equation (\ref{aug_lagrange1}) with respect to $\Theta_i$ corresponds to setting:

\begin{equation}
\label{step1}
\Theta_i^j =  \underset{\Theta_i}{\mbox{argmin}} \left \{ -\mbox{log det } \Theta_i + \mbox{trace} (S_i \Theta_i)+ \nicefrac{\gamma}{2} || \Theta_i - Z_i^{j-1} + U_i^{j-1} ||^2_2
\right \}.
\end{equation}

From equation (\ref{step1}) we can further understand the process occurring at this step. If $\gamma$ is set to zero only the negative log-likelihood 
terms will be left in equation (\ref{step1}) resulting in $\Theta_i^j = S_i^{-1}$, the maximum likelihood estimator. However, this will not
enforce either sparsity or temporal homogeneity and requires the assumption that $S_i$ is invertible. Setting $\gamma$ to be a positive constant
implies that $\Theta_i$ will be a compromise between minimising the negative log-likelihood and remaining in the proximity of $Z_i^{j-1}$.
The extent to which the latter is enforced will be determined by both $\gamma$ and Lagrange multiplier $U_i^{j-1}$. As we will see in step 2, it is the
$\{Z_i\}$ terms which encode the sparsity and temporal homogeneity constraints.


Differentiating the right hand side of equation (\ref{step1}) with respect to $\Theta_i$
and setting the derivative equal to zero yields:
\begin{equation}
\Theta_i^{-1} - \gamma \Theta_i = S_i -\gamma \left (Z^{j-1}_i - U^{j-1}_i \right )
\end{equation}
which is a matrix quadratic in $\Theta_i$ (after multiplying through by $\Theta_i$).
In order to solve this quadratic, we observe that both
$\Theta_i$ and $S_i - \gamma \left (Z^{j-1}_i - U^{j-1}_i \right )$ share
the same eigenvectors (see Appendix \ref{app_2}). This allows us to solve equation (\ref{step1}) using an eigendecomposition 
as outlined below.
Now letting $\theta_r$ and $s_r$ denote the $r$th 
eigenvalues of $\Theta_i$ and $S_i - \gamma \left (Z^{j-1}_i - U^{j-1}_i \right )$ respectively we have that:
\begin{equation}
\label{quad}
\theta_r^{-1} - \gamma \theta_r = s_r .
\end{equation}
Solving the quadratic in equation (\ref{quad}) yields 
\begin{equation}
\theta_r = \frac{1}{2\gamma} \left ( -s_r + \sqrt{s_r^2 + 4 \gamma}\right ), 
\label{eval_equation}
\end{equation}
for $r=1,\ldots, p$. Due to the nature of equation (\ref{eval_equation}) it follows that all eigenvalues, $\theta_i$ will be great than zero.
Thus Step 1 involves an eigendecomposition and update
\begin{equation}
\Theta_i = V_i \tilde{D_i}V_i^T
\label{theta_update}
\end{equation}
for each $i= 1, \ldots, T$. Here 
$V_i$ is a matrix containing
the eigenvectors of $S_i - \gamma \left (Z^{j-1}_i - U^{j-1}_i \right )$ and $\tilde{D_i}$ is a diagonal matrix containing 
entries $\theta_1, \ldots, \theta_p$. As discussed, all of the entries in $\tilde{D_i}$ will be strictly positive, ensuring that
each $\Theta_i$ will be positive definite. Moreover, we also note from equation (\ref{theta_update}) that each $\Theta_i$ will
also be symmetric.

\subsubsection*{Step 2: Update $ \{Z^j_i\} $ }

As in step 1, all variables $\{\Theta^j_i\}$ and $\{U^j_i\}$ are treated as constants when updating $\{Z_i\}$ variables. Due to the 
presence of the Fused Lasso penalty in equation (\ref{aug_lagrange1}) we cannot update each $Z^j_i$ separately as was the case with each
$\Theta^j_i$ in step 1. 
Instead, at the $j$th iteration the $\{Z^j_i\}$ variables are updated by solving: 
\small
\begin{equation}
\label{step22}
 \{Z^j_i\} = \underset{\{Z_i\}}{\mbox{argmin}} \hspace{5mm}  \left \{
  \nicefrac{\gamma}{2}\sum_{i=1}^T || \Theta_i^j - Z_i + U_i^{j-1} ||_{2}^2 + \lambda_1 \sum_{i=1}^{T}||Z_i||_1 +\lambda_2 \sum_{i=2}^T ||Z_i - Z_{i-1}||_1  \right \},
\end{equation}
\normalsize
where we note that only element-wise operations are applied.
As a result it is possible to break down equation (\ref{step22})
into element-wise optimisations of the following form:
\small
\begin{equation}
\label{fused_objective_thing}
 \underset{\{Z_i\}_{k,l}}{\mbox{argmin}} \hspace{5mm}  \left \{   \nicefrac{\gamma}{2}\sum_{i=1}^T || (\Theta_i^j - Z_i + U_i^{j-1})_{k,l}||_{2}^2 
+ \lambda_1 \sum_{i=1}^{T}||(Z_i)_{k,l}||_1 +\lambda_2 \sum_{i=2}^T ||(Z_i - Z_{i-1})_{k,l}||_1 \right \}
\end{equation}
\normalsize
where we write $(M)_{k,l}$ to denote the $(k,l)$ entry for any square matrix $M$.
This corresponds to a 
 \textit{Fused Lasso signal approximator}  problem \citep{hoefling} (see Appendix \ref{app_5}). Moreover, due 
to the symmetric nature of matrices $\{\Theta_i\}$, $\{Z_i\}$ and $\{U_i\}$ we require $\frac{p(p+1)}{2}$ optimisations 
of the form shown in equation (\ref{fused_objective_thing}). Thus by introducing auxiliary variables
$\{Z_i\}$ and formulating the augmented Lagrangian we are able to enforce both the sparsity and temporal homogeneity
penalties by solving a series of one-dimensional Fused Lasso optimisations.

\subsubsection*{Step 3: Update $\{U_i^j\}$}

Step 3 corresponds to an update of Lagrange multipliers $\{U_i^j\}$ as follows:
\begin{equation}
 U_i^j = U_i^{j-1} + \Theta_i^j - Z_i^j \hspace{2mm} \mbox{for } i=1,\ldots, T
\end{equation}


\subsubsection{Convergence Criteria} 

The proposed algorithm is
 an iterative procedure consisting of Steps 1-3 described above until convergence is reached.
In order to guarantee convergence we require both primal and dual feasibility: primal feasibility 
refers to satisfying the constraint $\Theta_i=Z_i$ while dual feasibility refers to minimisation 
of the Augmented Lagrangian. That is we require both that $\nabla_{\Theta} \mathcal{L}(\Theta, Z^j, U^j)=0 $ and 
$ \nabla_{Z} \mathcal{L}(\Theta^{j+1}, Z, U^j)=0$. We can check for primal feasibility by considering $|| \Theta_i^j - Z_i^j||_2^2$ 
at each iteration. 
As detailed in Appendix \ref{app_4}, step 3 ensures that $\{Z_i\}$ are always dual feasible and it sufficies to consider
$||Z^{j}-Z^{j-1}||^2_2$ to verify dual feasibility in $\{\Theta_i\}$ variables. 
 Thus 
the SINGLE algorithm is said to converge when $|| \Theta_i^j - Z_i^j||_2^2 < \epsilon_1$ and $||Z_i^{j} - Z_i^{j-1}||_2^2 < \epsilon_2$ for $i=1,\ldots, T$ 
where $\epsilon_1$ and $\epsilon_2$ are user specified convergence thresholds. 
The complete procedure is given in Algorithm \ref{psuedo}. 

\begin{algorithm}[ht]
 \SetAlgoLined
 
  \textbf{Input:} Multivariate fMRI time series $X_1,...,X_T$, where each $X_i \in \mathbb{R}^{1 \times p}$, \\
  Gaussian kernel size 
  $h$, 
  penalty parameters $\lambda_1, \lambda_2$, 
  convergence tolerance $\epsilon_1, \epsilon_2$, \\
  max number of iterations $M$;\\
 \KwResult{Sparse estimates of precision matrices $  \Theta_1,..., \Theta_T$.}
Set $\Theta_i^0 = I_p$, $Z_i^0 = U_i^0 = \mbox{\boldmath$0$}$ for $  i \in \{1,...,T\}$ and $j=1$\;
\For{i in $\{1, \ldots, T\}$}{
$ \mu_i = \frac{ \sum_{j=1}^T K_h(i, j) \cdot X_j}{\sum_{j=1}^T K_h(i, j)}$\;
}
\For{i in $\{1, \ldots, T\}$}{
$S_i = \frac{ \sum_{j=1}^T K_h(i, j) \cdot (X_j-\mu_j)^T (X_j-\mu_j)}{\sum_{j=1}^T K_h(i, j)}$\;
}
\While{Convergence ==  False and $ j < M$}{
  \#\# $\{\Theta\}$ Update\;
  \For{i in $\{1, \ldots, T\}$}{
  $V,D = \mbox{eigen} \left (S_i - \gamma \left (Z_i^{j-1} - U_i^{j-1} \right ) \right )$\;
  $\tilde D = \mbox{diag} \left ( \frac{1}{2\gamma} \left ( -D + \sqrt{D^2 + 4\gamma} \right )\right )$\;
  $\Theta_i^j = V \tilde D V^T$\;
  }
  \#\# $\{Z\}$ Update\;
  \For{l in $\{1, \ldots, p \}$}{
  \For{k in $\{l, \ldots, p \}$}{
  $x = \mbox{concat}\left ( \left (\Theta_1^j - U_1^{j-1} \right )_{k,l}, \ldots, \left (\Theta_T^j - U_T^{j-1} \right )_{k,l}  \right )$\;
  $\hat Z = \mbox{FLSA} (x, \lambda_1, \lambda_2)$\;
  $\left (Z_1^j, \ldots, Z_T^j \right )_{k,l} = \hat Z $\;
  }
  }
  
  \#\# $\{U\}$ Update\;
  \For{i in $\{1, \ldots, T\}$}{
  $U_i^j = U_i^{j-1} + \Theta_i^j - Z_i^j$\;
  }
  \eIf{$||\Theta_i^j - Z_i^j||^2_2 < \epsilon_1 \mbox{ and } ||Z_i^{j}-Z_i^{j-1}||_2^2 < \epsilon_2, \hspace{2mm} \forall i$ }{
   $\mbox{Convergence=True}$\;
   }{
  j = j + 1\;
  }

}
\Return  $  \Theta_1,...,  \Theta_T$
 \caption{Smooth Incremental Graphical Lasso Estimation (SINGLE) algorithm}
 \label{psuedo}
\end{algorithm}

\subsubsection{Computational Complexity}
\label{sec::comp_complex}
As discussed previously the optimisation of the SINGLE objective function involves the iteration of three steps. In step 1 we 
perform $n$ eigendecompositions, each of complexity $\mathcal{O} (p^3)$
where $p$ is the number of nodes (i.e., the dimensionality of the data). Thus step 1 has a computational complexity of $\mathcal{O}(n p^3)$.
We note that step 2 requires $\frac{p(p+1)}{2}$ iterations of the Fused Lasso\footnote{$\frac{p(p-1)}{2}$ edges and $p$ more along the diagonal} where each 
iteration is $\mathcal{O}\left(n \mbox{log}( n)\right)$ \citep{hoefling}. Thus the computational complexity of step 2 is  $\mathcal{O}\left(p^2n \mbox{log}( n)\right)$.
Finally step 3 only involves matrix addition implying that the final computational complexity of the SINGLE algorithm is $\mathcal{O}\left(p^2n \mbox{log}( n) + np^3\right)$.
This is dominated by the number of nodes, $p$, not the number of observations. As a result the limiting factor is likely to
be the number of nodes in a study.

\subsection{Parameter Tuning} \label{param_select}

The SINGLE algorithm requires the input of three parameters which can be tuned using the available data: $\lambda_1, \lambda_2$ and $h$.
Each of these parameters has a direct interpretation. Parameter $h$ is the width of the Gaussian kernel. Following from our
discussions in section \ref{sec::cov_matrix_est}, similar considerations 
should be made when tuning $h$ as when tuning the width of a sliding window. Parameters $\lambda_1$ and $\lambda_2$ affect the sparsity and temporal
homogeneity respectively. In particular, increasing $\lambda_1$ will result in network estimates with a higher degree of sparsity whereas
increasing the value of $\lambda_2$ will encourage the fusion of temporally adjacent estimates. We discuss each of these three parameters in turn.

The choice of parameter $h$ describes certain assumptions relating to the nature of the available data which are often not 
formally discussed. 
The use of a kernel (be it in the form of a sliding window or otherwise)
also reflects an assumption of local, as opposed to global, stationarity.
This assumption is that it is possible to obtain time dependent parameter estimates that accurately reflect the correlation 
structure within a neighbourhood of any observation but possibly not over an arbitrarily long time horizon. The choice of $h$ 
can therefore be seen as an assumption relating to the extent of non-stationarity of the available data
(for an attempted definition of the degree of non-stationarity see \citet[chap. 16]{AF}).

On the one hand, the choice of a large value of $h$ is indicative of an assumption that the data is close to stationary. If this 
is the case, a large choice of $h$ allows for the accurate estimation of sample covariance matrices by incorporating information
 across a wide range of observations. However, if this assumption is incorrect, the choice of a large $h$ can result in overly 
smoothed estimates where short term variation is overlooked. On the other hand, the choice of a small $h$ implies an assumption of 
a higher degree of non-stationarity. Here
the choice of a small $h$ can allow for the accurate estimation of sample covariance matrices by correctly discarding irrelevant information. 
However reducing the value of $h$ will result in an increase in the variance of the estimators
as it implies that a smaller sample size is used to estimate parameters. This
effect is more dramatic for large values of $p$ as a greater number of parameters must be estimated. Overall, the best performing value of $h$ in any
given setting will depend on the difficulty of the estimation task, in particular the dimensionality of $p$, as well as the rate of change  
 of the 
underlying networks. The latter is not known apriori in many fMRI applications. 

To avoid making specific assumptions about the nature of the temporal variability we rely on an entirely data-driven technique
when choosing $h$ that best describes the observations. The approach taken here is to use cross-validation \citep{silverman1986density}.
As before, goodness-of-fit is employed to quantify how well estimated sample covariance matrices describe the observed time series. 
We define the leave-one-out (LOO) log-likelihood for the $i$th observation and some fixed choice of $h$ as follows:
\begin{equation}
\mathcal{L}_{-i}(h) =  -\frac{1}{2} \mbox{ log det }\left (S^{(h)}_{-i} \right) - \frac{1}{2} \left (X_i - \mu_{-i}^{(h)} \right )^T \left (S_{-i}^{(h)} \right )^{-1} \left (X_i - \mu_{-i}^{(h)} \right ),
\end{equation}
where both $\mu_{-i}^{(h)}$ 
and $S_{-i}^{(h)}$ are estimated with 
the $i$th observation removed for a given $h$. Thus $\mathcal{L}_{-i}(h)$ allows us to estimate the goodness-of-fit at $X_i$ for any fixed $h$.
 We subsequently choose $h$ in order to maximise the following score function:
\begin{equation}
\label{h_tune}
CV(h) = \sum_{i=1}^T \mathcal{L}_{-i}(h).
\end{equation}

Parameters $\lambda_1$ and $\lambda_2$ determine the sparsity and temporal homogeneity of the estimated networks
respectively. Therefore $\lambda_1$ and $\lambda_2$ directly affect the degrees of freedom of the estimated networks. 
In this case we can employ a more sophisticated parameter tuning technique based on the Akaike Information Criterion (AIC).
The use of AIC allows us to estimate the in-sample prediction error for each choice of
parameters $\lambda_1$ and $\lambda_2$, allowing for a clear
comparison across different values of each parameter \citep[chap. 7]{hastie2009elements}.
For any pair $\lambda_1, \lambda_2$ we define the AIC as:
\begin{equation}
AIC(\lambda_1, \lambda_2) = 2  \sum_{i=1}^T \left (  - \mbox{log det }(\Theta_i) + \mbox{ trace }(S_i \Theta_i) \right ) + 2 K
\end{equation}
where $K$ is the estimated degrees of freedom. 
For a given range of $\lambda_1$ 
and $\lambda_2$ values an extensive grid-search is performed
with the resulting choices of $\lambda_1$ and $\lambda_2$ being the pair that minimises $AIC$.

Following \cite{FusedLasso} we define $K$ to be the number of non-zero coefficient
blocks in $ \{ (\Theta_{i})_{r,s} \}_{i=1,\ldots, T}$ for $1 \leq r \neq s \leq p$. That is, we count a sequence of one or more consecutive non-zero 
and equal estimates of partial correlations as one degree of freedom. This can be formally written as:
\begin{equation}
K = \sum_{r,s} \sum_{i=2}^T \mathbbm{1} \left (  (\Theta_i)_{r,s} \neq (\Theta_{i-1})_{r,s} \cap (\Theta_i)_{r,s} \neq 0 \right ).
\end{equation}

\subsubsection{Comparison to Related Work}

There are currently limited methodologies available for estimating dynamic functional connectivity networks. A novel approach has recently been 
proposed in the form of the DCR algorithm \citep{DCR}. The DCR is able to estimate functional connectivity networks
by first partitioning time series into piece-wise stationary segments. This allows the DCR to exploit the vast literature relating to
stationary network estimation. Formally, the DCR algorithm detects statistically significant 
change points by applying a block bootstrap permutation test. The use 
of a block bootstrap allows the DCR algorithm to account for autocorrelation present in fMRI data. 

A common approach involves the use of a sliding window \citep{Hutchinson2013}. This involves recursively estimating covariance matrices $S_i$ by re-weighting 
observations according to a sliding window or kernel. Subsequently, analysis can be performed directly on $S_i$ to infer the network
structure at the $i$th observation.
This approach is studied in detail by \cite{wasserman}.
However sliding window approaches
face the potential issue of variability between temporally adjacent networks. This arises as a direct consequence of the fact that each network is
estimated independently without any mechanism present to encourage temporal homogeneity. 
We believe this additional variability can jeopardise the accuracy of the estimation procedure and can result in networks which do not
reflect the true network structure over time. 
The SINGLE algorithm addresses precisely this problem by introducing 
an additional Fused Lasso penalty. In this way, changes in the connectivity structure are only reported when strongly validated by the data.
The effect of the additional Fused Lasso penalty is studied extensively in the simulation study provided in section \ref{simulations}.

Finally, the SINGLE algorithm is formally related to the Joint Graphical Lasso (JGL) \citep{JGL}.
The JGL was designed with the motivation of improving network inference by leveraging information across related observations and data sets. 
However, while the JGL focuses on stationary network estimation the SINGLE algorithm is designed to estimate dynamic networks. This manifests itself in 
two main differences to the overall objective functions of each of the algorithms. Firstly, the SINGLE algorithm only employs the Fused Lasso
penalty as the Group Lasso penalty proposed in \cite{JGL} cannot be used in the context of temporal homogeneity. This is due to the fact that
the Group Lasso penalty encourages all coefficients to
either be zero or non-zero in unison and therefore ignores temporal behaviour. Secondly, while both algorithms contain a Fused Lasso penalty the nature of these penalties are vastly different. In the
case of the JGL there is no natural ordering to observations and therefore \textit{fusions} are present between all networks (i.e., the penalty
is of the form $\sum_{i \neq j} || \Theta_i - \Theta_j||_1 $). This is not the case in the SINGLE algorithm where there is a chronological ordering. This results
in a penalty of the form $ \sum_{i=2}^T || \Theta_i - \Theta_{i-1}||_1$. 

\subsubsection*{Software}
The SINGLE algorithm is freely available as an R package, and can be downloaded along with its documentation from the Comprehensive 
R Archive Network (CRAN) \citep{SINGLE_vignette}.

\subsection{Experimental Data}
\label{sec::exp_data}

The data was collected from 24 healthy subjects performing a simple but attentionally demanding cognitive task. 
This fMRI data set is particularly challenging as the BOLD time series has a length of 126  and corresponds to 18 
ROIs, implying a low ratio of observations to dimensionality. We expect there to be a change in correlation 
structure approximately every 15 time points. Thus the number of observations available to estimate each connectivity structure is small 
relative to the number of ROIs.

In the CRT task, 24 subjects were presented with an initial fixation cross for 350ms. This was followed by a response cue in
the form of an arrow in the direction of the required response and lasting 1400ms. The inter-stimulus interval
was 1750ms. Finger-press responses were made with the index finger of each hand. Subjects were instructed to respond as quickly 
and as accurately as possible. To maximise design efficiency, stimulus presentation was blocked, with five repeated blocks of 14 
response trials interlaced with five blocks of 14 rest trials, and four response trials at the start of the experiment. This 
resulted in a total of 74 response trials per subject.

Image pre-processing involved realignment of EPI images to remove the effects of motion between scans, spatial smoothing using a 6mm
full-width half-maximum Gaussian kernel, pre-whitening using FILM and temporal high-pass filtering using a cut-off frequency of $\nicefrac{1}{50}$
Hz to correct for baseline drifts in the signal. FMRIB's Linear Image Registration Tool (FLIRT) \citep{Smith2004} was used to register EPI functional 
data sets into standard MNI space using the participant's individual high-resolution anatomical images.

The nodes were eighteen cortical spherical regions based on \cite{pandit2013traumatic}. Briefly, these nodes were defined based on peak regions 
from a spatial group independent components analysis of resting state fMRI. The regions were chosen for 
the nodes to encompass a wide range of cortical regions including regions within two well recognised functional connectivity networks, the fronto-parietal 
cognitive control network (FPCN) and default mode network (DMN) regions, as well as motor, visual and auditory cortical regions (see Table \ref{table:MNI}). 
For each subject and node the mean time-course from within a 10mm diameter sphere centred on each of the 18 peaks was calculated. Six motion parameters,
estimated during realignment, were filtered out of each time-course, using linear regression. The resulting 18 time-courses were subsequently used. 

\begin{table}[ht]
\centering 
\begin{tabular}{r l l l l} 
\hline\hline 
Number  & Name  &  \multicolumn{3}{c} {MNI} coordinates (mm) \\ 
\hline 
1 & Left Lateral Parietal (DMN)	 & 	-46 & 	-62 & 	24\\
2 &  Right Lateral Parietal (DMN)	 & 	50	 & -54 & 	16\\
3 &  Posterior Cingulate Cortex (DMN) & 		-2	 & -46 & 	20\\
4 &  Ventromedial PFC (DMN)	 & 	2 & 	54 & 	8\\
5  & Ventromedial PFC (FPCN)	 & 	-2	 & 54	 & 20\\
6  & Dorsal Anterior Cingulate/preSMA (FPCN)	 & 	2	 & 26 & 	56\\
7  & Left Inferior Frontal Gyrus (FPCN)	 & 	-46 & 	22 & 	-12\\
8  & Right Inferior Fronal Gyrus (FPCN)	 & 	54 & 	22 & 	-4\\
9 &  Left Inferior Parietal (FPCN)	 & 	-54	 & -54 & 	20\\
10 &  Right Inferior Parietal (FPCN) 	 & 	54	 & -54	 & 16\\
11  & Left Superior Temporal Sulcus (FPCN) & 		-50	 & -26 & 	-12\\
12  & Right Superior Temporal Sulcus (FPCN) & 		54 & 	-22 & 	-12\\
13  & Posterior Cingulate Cortex (FPCN)	 & 	-2	 & -50 & 	24\\
14  & Left Motor	 & 	-38 & 	-22	 & 52\\
15  & Left Primary Auditory	 & 	-54	 & -18 & 	0\\
16  & Primary Visual		 & 2 & 	-74 & 	4\\
17  & Right Motor		 & 34	 & -22	 & 52\\
18  & Right Primary Auditory		 & 62 & 	-18 & 	8\\
\hline 
\end{tabular}
\caption{Regions and MNI coordinates} 
\label{table:MNI} 
\end{table}

\section{Experimental Results}
\label{simulations}

\subsection{Simulation settings}

In this section we evaluate the performance of the SINGLE algorithm through a series of simulation studies.
In each simulation we produce simulated time series data giving rise to a number of 
connectivity patterns which reflect those reported in real fMRI data. The objective is then to measure
whether our proposed algorithm is able recover the underlying patterns.
That is, we are interested primarily in the correct estimation of the presence or absence of
edges.  

There are two main properties of fMRI data which we wish to recreate in the simulation study. The first is the high autocorrelation
which is typically present in fMRI data \citep{handbook}. The second and main property we wish to recreate
is the structure of the connectivity networks themselves. It is widely reported that brain networks have a small-world topology as well as 
highly connected hub nodes \citep{bullmore} and we therefore look to enforce these properties in our simulations.

Vector Autoregressive (VAR) processes are well suited to the task of producing autocorrelated multivariate time series as they are capable of
encoding 
autocorrelations within components as well as cross correlations
across components \citep{DCR}. 
Moreover, when simulating connectivity structures we study the performance of the proposed algorithm using three types of random graphs;
Erd\H{o}s-R\'{e}nyi random graphs \citep{ErdosRenyi}, scale-free random graphs obtained by using
the preferential attachment model of \cite{barabasi1999emergence}
and small-world random graphs obtained using the \cite{watts1998collective} model.
Erd\H{o}s-R\'{e}nyi random graphs are included as they correspond to the simplest and most widely studied type of random network while
the use of scale-free and small-world networks is motivated by the fact that they are each known to each resemble different aspects of fMRI networks.

When simulating Erd\H{o}s-R\'{e}nyi random networks we maintain the edge strength of the connectivity between nodes fixed at 0.6. However, when simulating 
scale-free and small-world networks we randomly sample the edge strengths uniformly from
$[-\nicefrac{1}{2},-\nicefrac{1}{4}] \cup [\nicefrac{1}{4}, \nicefrac{1}{2}]$. 
This additional variability in the edge strength together with the reduced expected magnitude of each edge further 
increases the difficulty of the estimation task.

Each of the simulations considered in this section is aimed at studying the performance of the proposed algorithm in a different scenario. 
We begin by considering the overall performance of the SINGLE algorithm by generating connectivity structures according to 
Erd\H{o}s-R\'{e}nyi, scale-free and small-world networks in simulations \rom{1}a, \rom{1}b and \rom{1}c respectively.
In many task-based experiments it is the case that the task is repeated several times, thus we expect there to be cyclic behaviour within 
the true functional connectivity structure (i.e., connectivity alternates between two structures) and we study this scenario in simulations 
\rom{2}a, \rom{2}b and \rom{2}c.
In simulation \rom{3} we study the performance of the algorithm as the ratio of observations, $n$, to nodes, $p$, decreases. This simulation is 
critical as it is often the case that there is a low ratio of observations to nodes, especially when considering subject specific fMRI data.
In simulation \rom{4} we quantify the computational cost of the SINGLE algorithm. 
Throughout each of these simulations we benchmark the performance of the SINGLE algorithm against both the DCR algorithm and two sliding window 
based algorithms. Here a sliding window is employed to obtain time-dependent estimates of the sample covariance matrices and the Graphical Lasso
is subsequently used to estimate a sparse connectivity structure. 
In order to ensure a fair comparison, the sliding window approach is employed using both a uniform kernel as well as a
Gaussian kernel. A summary of all simulations can be found in table [\ref{simulation_table}].

\begin{table}[ht]
\centering
  \begin{tabular}{ |l | l | l |l |p {5.5cm}|}
    \hline
    Simulation & Network & Interval length & Edge strength& Properties \& Motivation\\ \hline \hline
    \rom{1}a &  Erd\H{o}s-R\'{e}nyi & $n$=100 & 0.6 &Simplest and most widely used random network\\ \hline
    \rom{1}b &  Scale-free &  $n$=100 & $[-\nicefrac{1}{2},-\nicefrac{1}{4}] \cup [\nicefrac{1}{4}, \nicefrac{1}{2}]$& Networks with highly connected hub nodes present \\ \hhline{-----}
    \rom{1}c &  Small-world &  $n$=100 & $[-\nicefrac{1}{2},-\nicefrac{1}{4}] \cup [\nicefrac{1}{4}, \nicefrac{1}{2}]$& Networks with small-world topology and high local clustering\\ \hline
    \rom{2}a &   Erd\H{o}s-R\'{e}nyi & $n$=100 & 0.6& Cyclic network structure which\\ \hhline{----~}
    \rom{2}b &   Scale-free & $n$=100 & $[-\nicefrac{1}{2},-\nicefrac{1}{4}] \cup [\nicefrac{1}{4}, \nicefrac{1}{2}]$&  may be expected in task-based \\ \hhline{----~}
    \rom{2}c &   Small-world & $n$=100 & $[-\nicefrac{1}{2},-\nicefrac{1}{4}] \cup [\nicefrac{1}{4}, \nicefrac{1}{2}]$ &fMRI studies\\ \hline    
    \rom{3}a &  Scale-free & $n \in \{10, \ldots, 100\}$ &$[(-\nicefrac{1}{2},-\nicefrac{1}{4}] \cup [\nicefrac{1}{4}, \nicefrac{1}{2}]$ & The ratio of observations, $n$, to the number of ROIs, $p$, is  \\ \hhline{----~}
    \rom{3}b &  Small-world & $n \in \{10, \ldots, 100\}$ &$[(-\nicefrac{1}{2},-\nicefrac{1}{4}] \cup [\nicefrac{1}{4}, \nicefrac{1}{2}]$ & decreased in order to study performed in the presence of rapid changes \\
     \hline
  \end{tabular}
  \caption{A summary of simulation settings and motivation behind each simulation. }
   \label{simulation_table}
\end{table}

\subsection{Performance measures}

We are primarily interested in the estimation of the functional connectivity graphs at every time point. 
In our setting this corresponds to correctly identifying the non-zero entries in estimated precision
matrices, $\hat \Theta_i$, at each $i=1, \ldots, T$. An edge is assumed to be present between the $j$th and $k$th nodes 
if $( \hat \Theta_i)_{j,k} \neq 0$. At the $i$th observation we define the set of all reported edges as $D_i = \{ (j,k): (\hat \Theta_i)_{j,k} \neq 0\}$.
We define the corresponding set of true edges as $T_i= \{ (j,k): ( \Theta_i)_{j,k} \neq 0\}$ where we write $\Theta_i$ to denote the 
true precision matrix at the $i$th observation.
Given $D_i$ and $T_i$ we consider a number of
performance measures at each observation $i$.

First we measure the precision, $P_i$. This measures the percentage of reported
edges which are actually present (i.e., true edges). Formally, the precision is given by:
$$ P_i =  \frac{|D_i \cap T_i|}{|D_i|}.$$
Second we also calculate 
the recall, $R_i$, formally defined as:
$$ R_i =  \frac{|D_i \cap T_i|}{|T_i|}. $$
This measures the percentage of true edges which were reported by each algorithm.
Ideally we would
like to have both precision and recall as close to one as possible. Finally, the $F_i$ \textit{score}, defined as
\begin{equation}
\label{f_eq}
F_i = 2  \frac{P_i  R_i}{P_i + R_i},
\end{equation}
summarises both the precision and recall by taking their harmonic mean. 



\subsubsection{Simulation \rom{1}a - Erd\H{o}s-R\'{e}nyi random networks}

In order to obtain a general overview of the performance of the SINGLE algorithm we simulate
data sets with the following structure: each data set consists of 3 segments each of length 100 (i.e., overall duration of 300). 
The correlation structure for each segment was randomly generated using an Erd\H{o}s-R\'{e}nyi random graph.
Finally a VAR process for each corresponding correlation
structure was simulated. Thus each data set consists of 2 change points at times $t=100$ and 200 respectively resulting in 
a network structure that is piece-wise constant over time. For this simulation the random graphs
were generated with 10 nodes and the probability of an edge between two nodes was fixed at $\theta=0.1$ (i.e., the expected number of edges is
$\theta \frac{p(p-1)}{2}$).

In the case of the SINGLE algorithm the value 
of $h$ was estimated by maximising the leave-one-out log-likelihood given in equation (\ref{h_tune}).
Values of $\lambda_1$ and $\lambda_2$ were estimated by minimising AIC. For the DCR algorithm, the block size for the 
block bootstrap permutation tests to be 15 and one thousand permutations where used for each permutation test.
In the case of the
sliding window and Gaussian kernel algorithms
the kernel width was estimated using leave-one-out log-likelihood and $\lambda_1$ was estimated by minimising AIC.

\begin{figure}[ht]
\centering
\includegraphics[width=\textwidth]{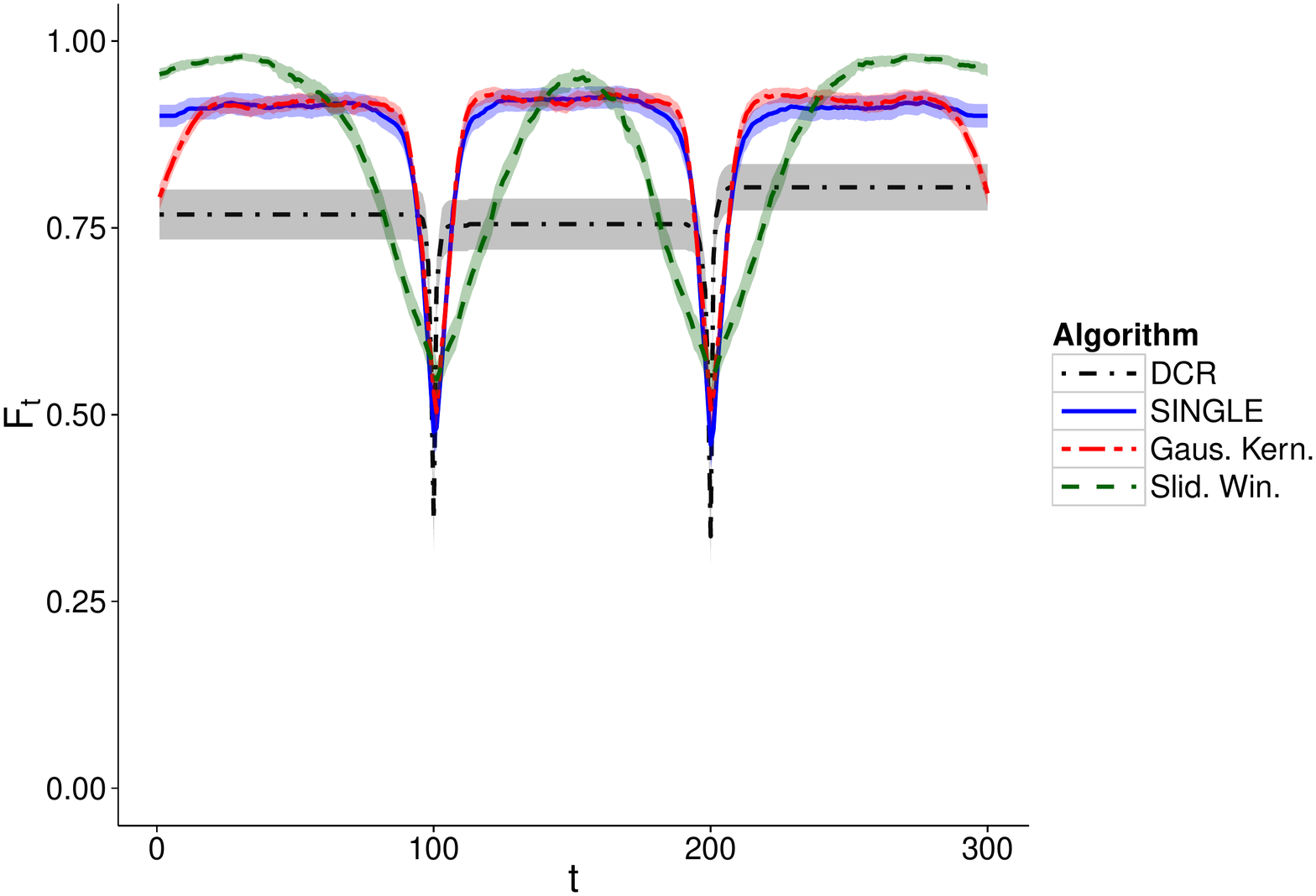}
\caption{Mean $F$ scores for Simulation \rom{1}a (shaded regions represent 95\% confidence intervals).
Here the underlying network structure was simulated using
Erd\H{o}s-R\'{e}nyi random networks and a change occurred every 100 time points. We note that all four algorithms experience a drop
in performance in the vicinity of these change points.}
\label{sim1a}
\end{figure}

In Figure [\ref{sim1a}] shows the average $F_t$ scores for each of the four algorithms over 500 simulations. We can see that the SINGLE algorithm 
performs competitively relative to the other algorithms. Specifically we note that the performance of the SINGLE algorithm mimics 
that of the Gaussian kernel algorithm.
We also note that all four algorithms experience a dramatic drop in performance in the vicinity of change points. This effect is most pronounced
for the sliding 
window algorithm.

\subsubsection{Simulation \rom{1}b - Scale-free networks}

It has been reported that brain networks are scale-free, implying that their degree distribution 
follows a power law. From a biological perspective this implies that there are a small but finite number of hub regions which have 
access to most other regions \citep{eguiluz2005scale}. While Erd\H{o}s-R\'{e}nyi random graphs offer a simple and 
powerful model from which to simulate random networks they fail to generate networks where 
the degree distribution follows a power law. In this simulation we analyse the performance of the SINGLE algorithm 
by simulating random networks according to the \cite{barabasi1999emergence} preferential attachment model. 
Here the power of preferential attachment was set to one.
Additionally, edge strength was also simulated according to a uniform distribution on 
$[-\nicefrac{1}{2},-\nicefrac{1}{4}] \cup [\nicefrac{1}{4},\nicefrac{1}{2}]$,
introducing further variability in the estimated 
networks.

\begin{figure}[ht]
\centering
\includegraphics[width=\textwidth]{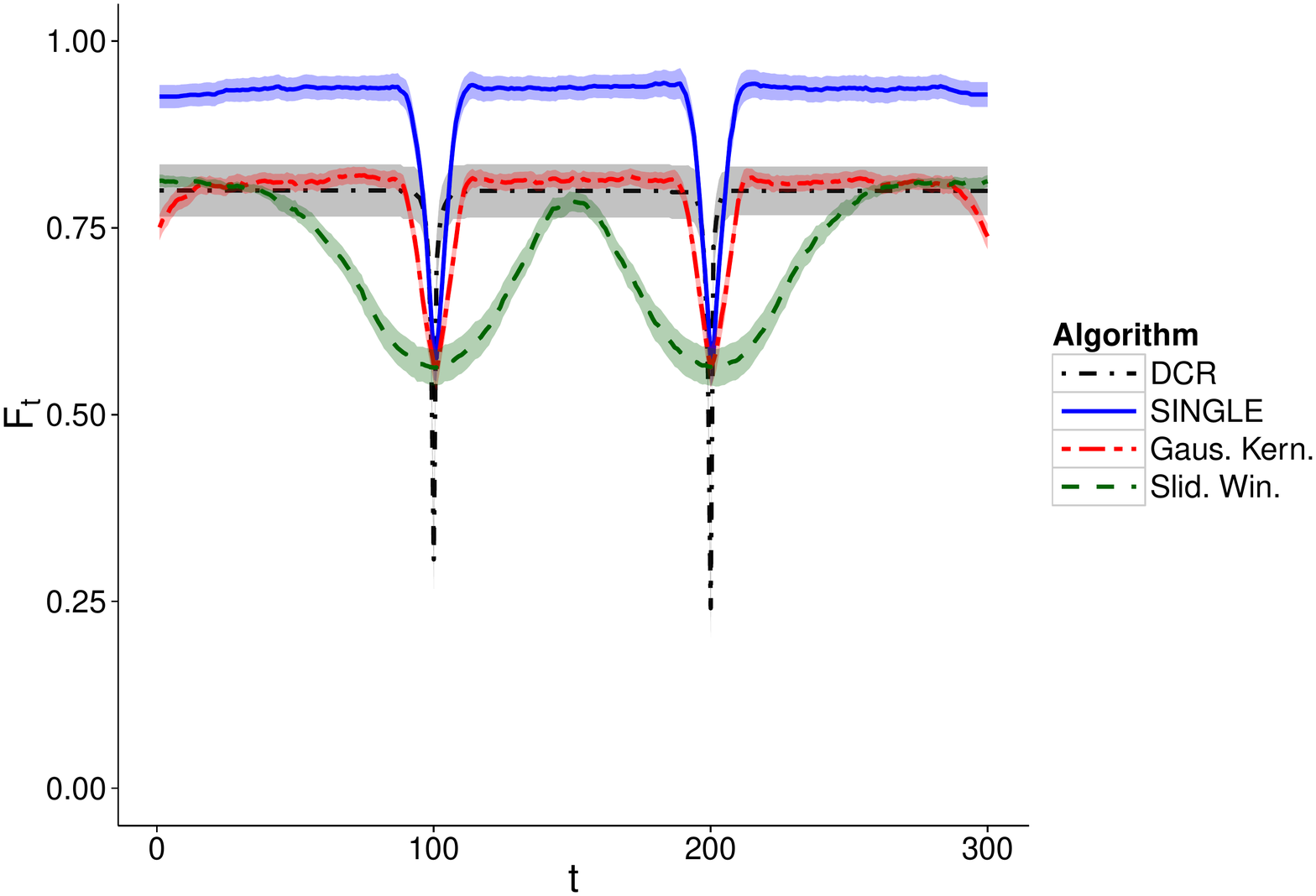}
\caption{
Mean $F$ scores for Simulation \rom{1}b (shaded regions represent 95\% confidence intervals).
Here the underlying network structure was simulated using
scale-free random networks according to the preferential attachment model of \cite{barabasi1999emergence}. 
A change occurred every 100 time points. We note that all four algorithms experience a drop
in performance in the vicinity of these change points. A full description of 
simulations is provided in Table [\ref{simulation_table}].}
\label{sim1b}
\end{figure}

In Figure [\ref{sim1b}] we see the average $F_t$ scores for each of the four algorithms over 500 simulations. We note that the performance of
the SINGLE and DCR
algorithms is largely unaffected by the increased complexity of the simulation. This is not true in the case of the sliding window and Gaussian 
kernel algorithms, both of 
which see their performance drop. We attribute this drop in performance to the fall in the signal-to-noise ratio and to the increased
complexity of the network structure. Similar results 
confirming that networks with skewed degree distributions (e.g., power-law distributions)
are typically harder to estimate have also been described in \cite{peng2009partial}. Finally,


\subsubsection{Simulation \rom{1}c - Small-world networks}

It has been widely reported that brain networks have a small-world topology \citep{stephan2000computational, sporns2004organization, bassett2006small}. 
In this simulation, multivariate time series were simulated such that the correlation structure follows a
small-world graph according to the Watts-Strogatz model 
\citep{watts1998collective}. 
Starting with a regular lattice, this model is parameterised by $\beta \in [0,1]$ which quantifies the probability of randomly 
rewiring an edge. This results
in networks where there is a tendency for nodes to form clusters, formally referred to as
a high clustering coefficient. Both anatomical \citep{sporns2004organization} as well as the functional brain networks 
have been reported as exhibiting such a network topology \citep{bassett2006small}. 
Throughout this simulation we set $\beta=\nicefrac{3}{4}$ and 
edge strength was simulated according to a Uniform distribution on 
$[-\nicefrac{1}{2},-\nicefrac{1}{4}] \cup [\nicefrac{1}{4},\nicefrac{1}{2}]$.

\begin{figure}[ht]
\centering
\includegraphics[width=\textwidth]{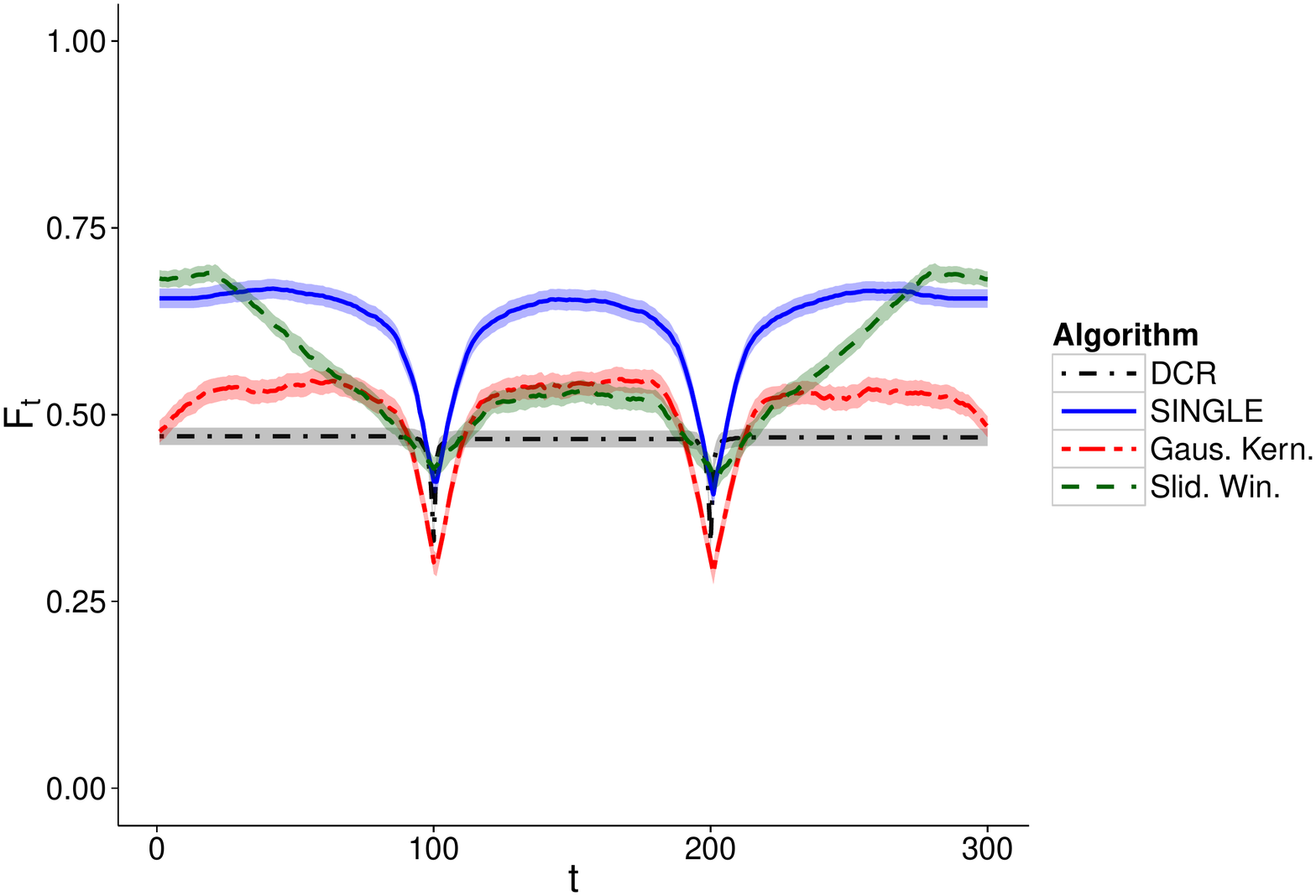}
\caption{
Mean $F$ scores for Simulation \rom{1}c (shaded regions represent 95\% confidence intervals).
Here the underlying network structure was simulated using
small-world random networks according to the Watts-Strogatz model. 
A change occurred every 100 time points. We note that all four algorithms experience a drop
in performance in the vicinity of these change points. A full description of 
simulations is provided in Table [\ref{simulation_table}].}
\label{sim1c}
\end{figure}

In Figure [\ref{sim1c}] we see the average $F_t$ scores for each of the four algorithms over 500 simulations. 
We note that there is a clear drop in the performance of all the algorithms relative to their performance
in simulations \rom{1}a and \rom{1}b. We believe this is due to the increased
complexity of small-world networks compared
to the previous networks we had considered. Formally, due to the high local clustering present in small-world networks,
the path length between any two nodes is relatively short.
As a result, we expect there to be a large number of correlated variables that are not directly connected. It has been reported
that the Lasso (and
therefore by extension the Graphical Lasso) cannot guarantee consistent
variable selection in the presence of highly correlated predictors \citep{zou2005regularization, zou2006adaptive}. Since all four algorithms are related to the 
Graphical Lasso, we conjecture that this may be the cause of the overall drop in performance.


\subsubsection{Simulation \rom{2}a - Cyclic Erd\H{o}s-R\'{e}nyi networks}

In task related experiments subjects are typically asked to alternate between performing a cognitive task and resting. 
As a result, we expect the functional connectivity structure to alternate between two states: a task related state and the resting state.
In order to recreate this scenario,
network structures are simulated in a cyclic
fashion such that the first and third correlation structures are identical.

\begin{figure}[ht]
\centering
\includegraphics[width=\textwidth]{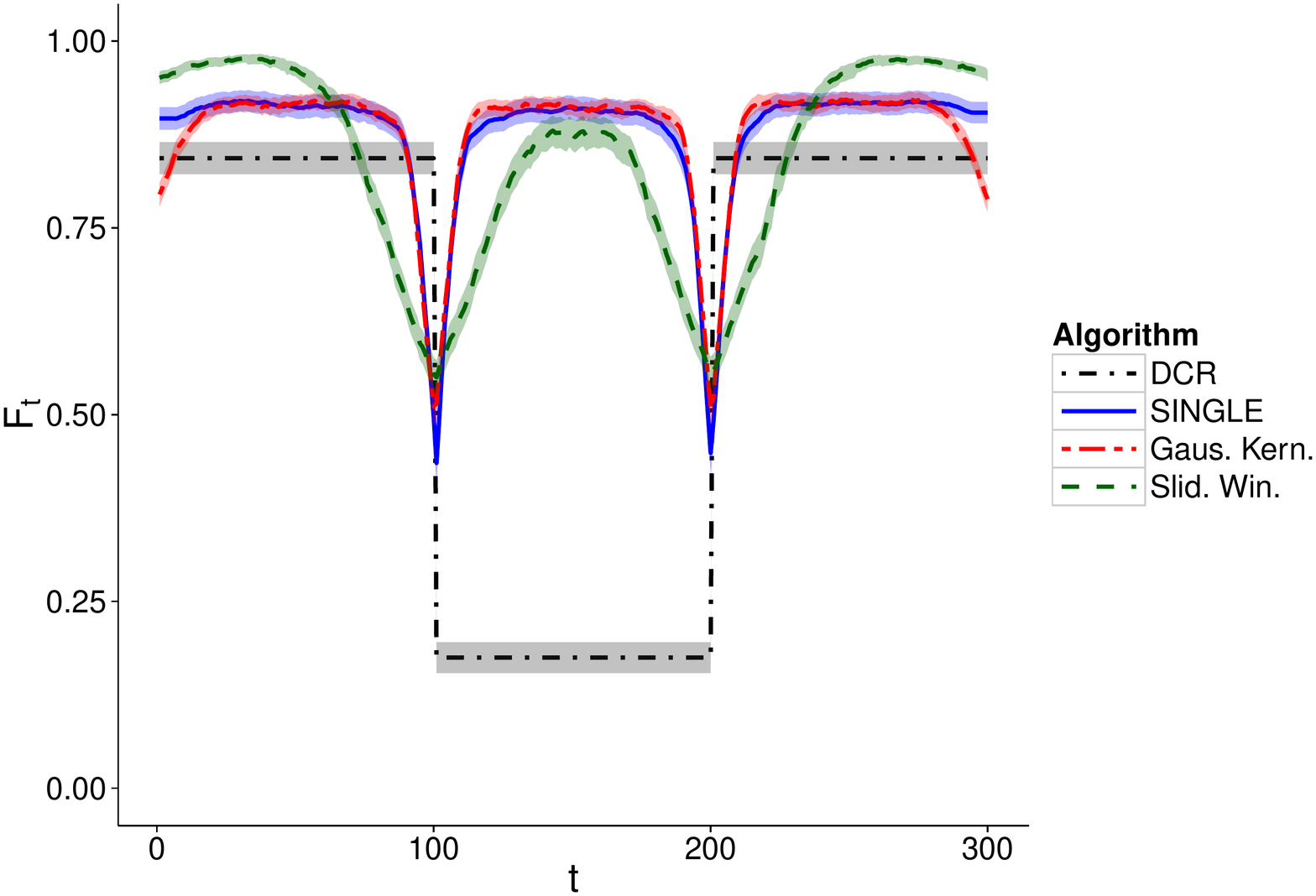}
\caption{
Mean $F$ scores for Simulation \rom{2}a (shaded regions represent 95\% confidence intervals).
Here the underlying network structure was simulated using
Erd\H{o}s-R\'{e}nyi random networks with the additional constraint that the first and third correlation structure be identical.
This gives rise to cyclic correlation structures which may be present in task-based studies.}
\label{sim2a}
\end{figure}

We note from Figure [\ref{sim2a}] that the performance of the SINGLE sliding window and Gaussian kernel algorithms is largely unaffected. 
However the DCR algorithm suffers
a clear drop in performance relative to simulation \rom{1}a. 
The drop in performance of the DCR algorithm is partly due to 
the presence of the recurring correlation structure. More specifically, we believe the problem to be related to 
the use of block bootstrapping permutation test to determine the significance of change points in the DCR. 
This test assumes that \textit{local} data points are correlated but expects data points that are \textit{far away}
to be independent. Typically this assumption holds. However when there is a recurring correlation structure, 
points that are far away may follow the same underlying distribution. As a result the power of the permutation test is heavily reduced.


\subsubsection{Simulation \rom{2}b - Cyclic Scale-free networks }

\begin{figure}[ht]
\centering
\includegraphics[width=\textwidth]{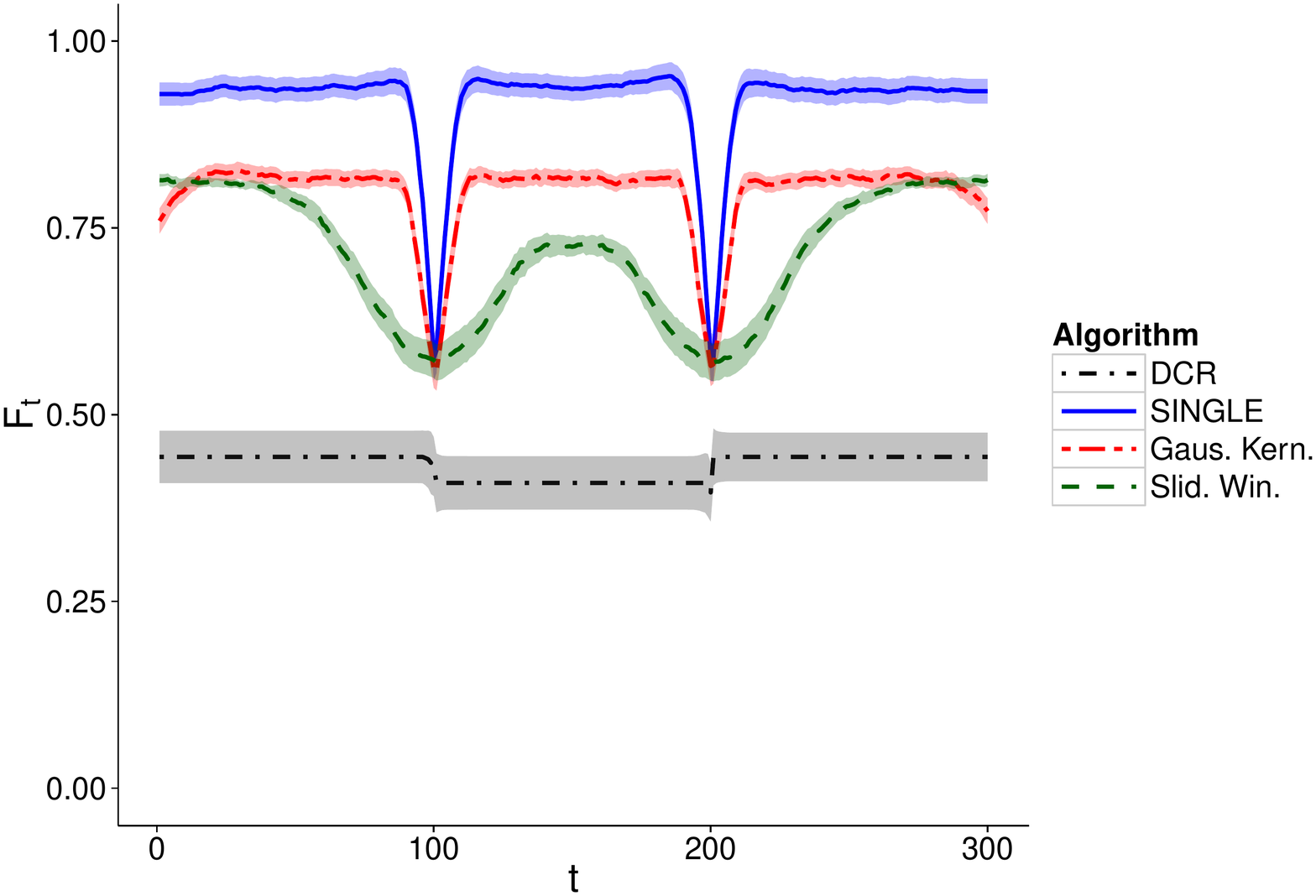}
\caption{
Mean $F$ scores for Simulation \rom{2}b (shaded regions represent 95\% confidence intervals).
Here the underlying network structure was simulated using
scale-free random networks with the additional constraint that the first and third correlation structure be identical.
This gives rise to cyclic correlation structures which may be present in task-based studies.}
\label{sim2b}
\end{figure}

In this simulation we simulate multivariate time series where the underlying correlation structure is cyclic and follows a scale-free distribution.
The results are summarised in Figures [\ref{sim2b}]. As in simulation \rom{1}b there is no noticeable difference in the 
performance of the SINGLE algorithm.
There is however a drop in the performance of the sliding window, Gaussian kernel and DCR algorithms.
This is particularly evident in the case of the DCR algorithm. As mentioned
previously the drop in performance of the sliding window and Gaussian kernel algorithms is due to the 
increased complexity of the network structure as well as the fall in the
signal to noise ratio. In the case of the DCR the drop in performance can be partly explained by the 
fact the assumptions behind the use of the block bootstrap no longer hold (see simulation \rom{2}a for a discussion) 
and the increased complexity of the network structure. These two factors combine to
greatly affect the performance of the DCR algorithm.


\subsubsection{Simulation \rom{2}c - Cyclic Small-world networks }

In this simulation we look to assess the performance of the SINGLE algorithm in a scenario that is representative of the 
experimental data we use in this work. As described previously, the experimental data used in this study corresponds to fMRI data from a Choice Reaction
Time (CRT) task. Here subjects are required to make rapid visually-cued motor decisions. Stimulus was presented in five on-task blocks each preceding 
a period where subjects were at rest. As a result we expect there to be a cyclic network structure. 

Thus in this simulation network structures are simulated in a cyclic fashion where each network structure is simulated according to a small-world 
network as in Simulation \rom{1}c. This simulation gives us a clear insight into the performance of the SINGLE algorithm in a scenario that is very 
similar to that proposed in the experimental data.

\begin{figure}[ht]
\centering
\includegraphics[width=\textwidth]{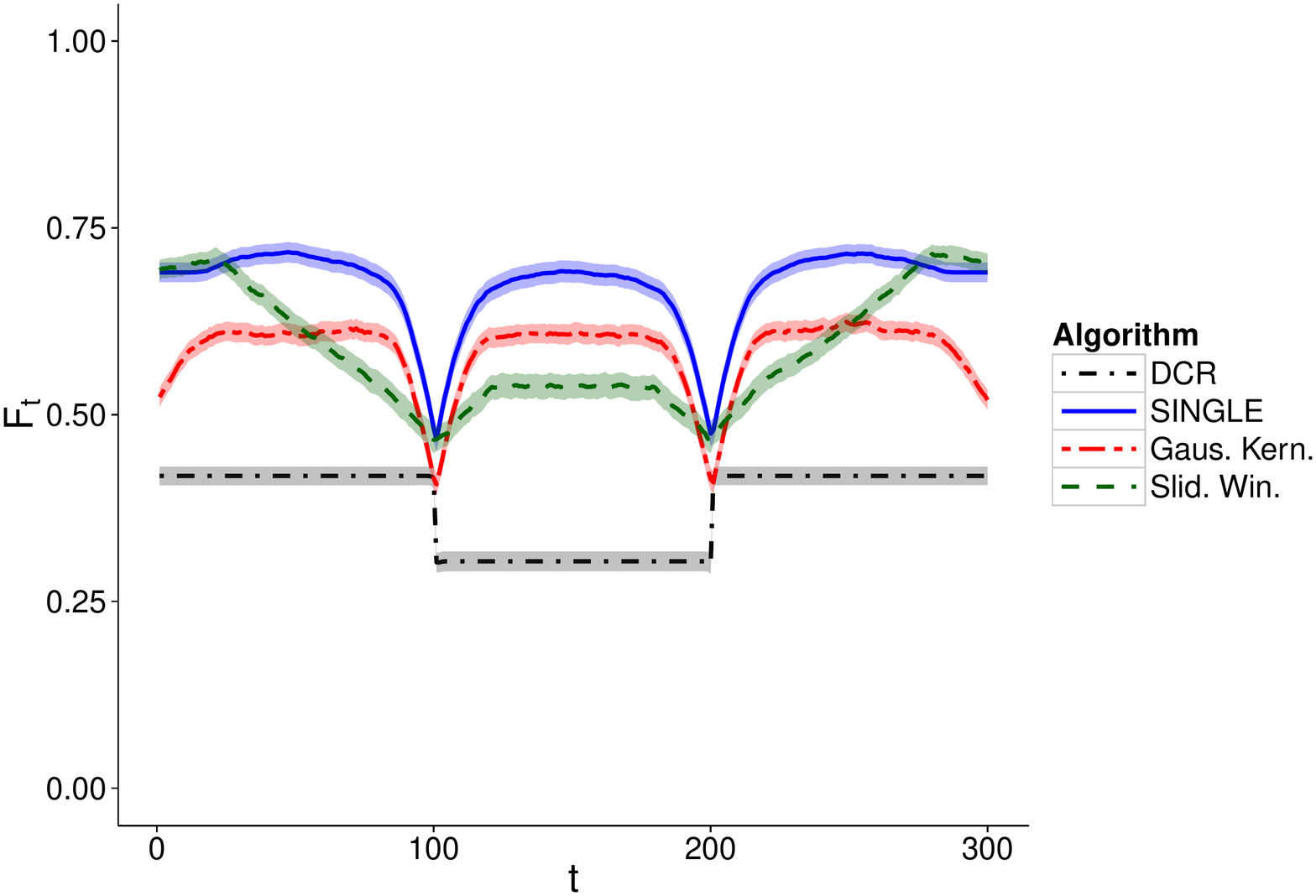}
\caption{
Mean $F$ scores for Simulation \rom{2}c (shaded regions represent 95\% confidence intervals).
Here the underlying network structure was simulated using
small-world random networks with the additional constraint that the first and third correlation structure be identical.
This gives rise to cyclic correlation structures which may be present in task-based studies.
}
\label{sim2c}
\end{figure}

The results are summarised in Figure [\ref{sim2c}]. We note that as in Simulation \rom{1}c there is drop in the performance of all four algorithms
relative to their performance in simulations \rom{2}a and \rom{2}b.
We believe this is due to the increased complexity of the underlying networks structures, specifically the high levels of clustering we experience 
in small-world networks which are not seen in Erd\H{o}s-R\'{e}nyi or scale-free random networks. 

\subsubsection{Simulation \rom{3}a - Scale-free networks with decreasing $\nicefrac{n}{p}$ ratio}

Here we study the behaviour of the proposed algorithm as the 
ratio of observations, $n$, to the number
of nodes, $p$, decreases. This is a particularly relevant problem in the case of fMRI data as it is often the case that the number of nodes in 
the study (typically the number of ROIs) will be much larger than the number of observations.

In this simulation we fix $p=10$ and allow the value of $n$ to decrease. Here we simulate a data set with three segments each of length $n$ where
the connectivity structure within each segment is randomly simulated according to a small-world network. Thus as the value of $n$ decreases we are able
to quantify the performance of the SINGLE algorithm in the presence of rapid changes in network structure. 

In the case of the SINGLE, sliding window and Gaussian kernel algorithms all parameters are estimated as discussed previously. In the case of the DCR algorithm
the value of block sizes for the block bootstrap test was also reduced accordingly.

\begin{figure}[ht]
  \centering
    \includegraphics[width=\textwidth]{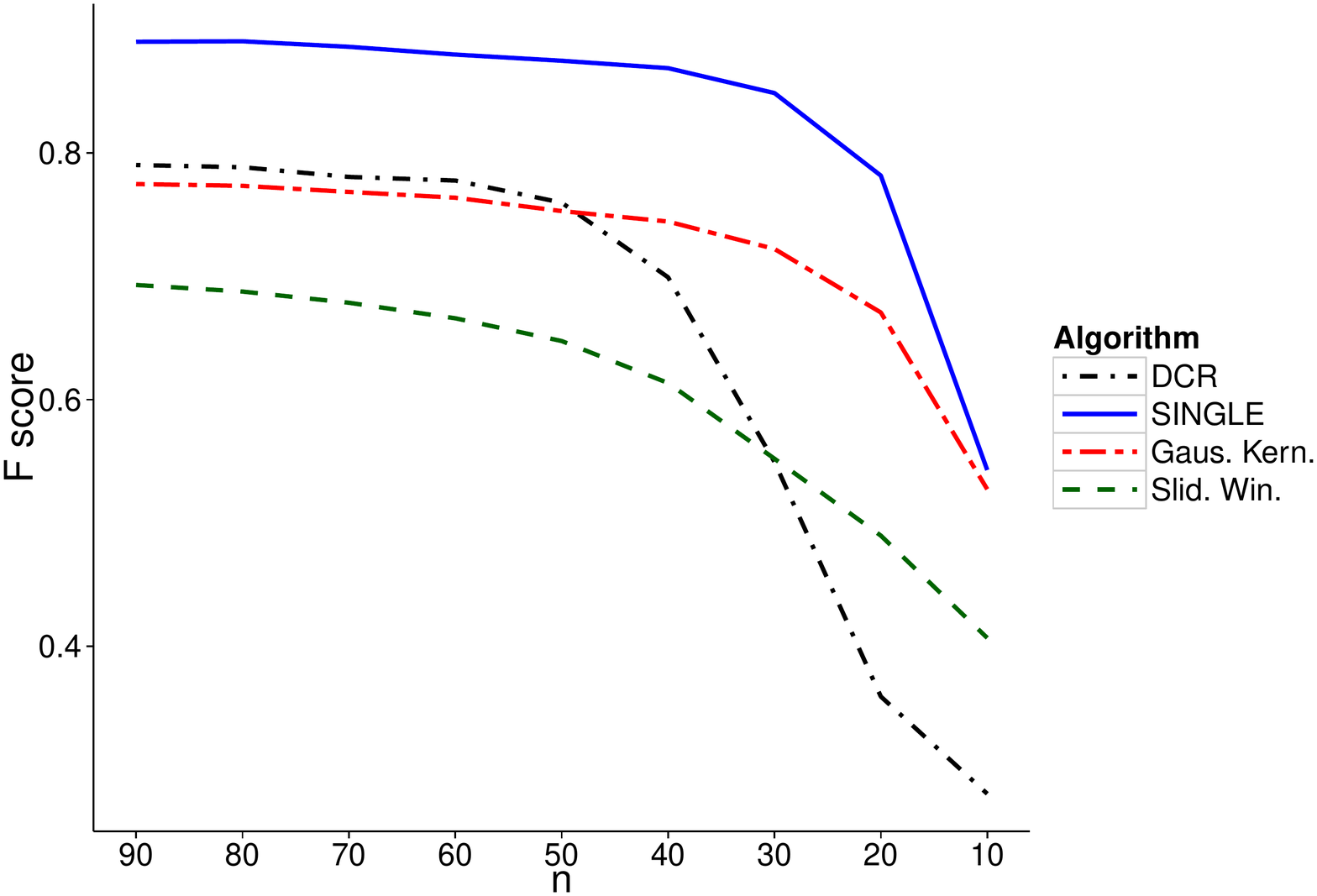}
      \caption{Results for Simulation \rom{3}a for the SINGLE, DCR and sliding window algorithms respectively over 500 simulations. 
      Here networks were simulated using scale-free random networks and the performance of each algorithm was studied as the ratio
      of observations, $n$, to the numbre of nodes, $p$, decreased. Here $p=10$ was fixed as $n$ decreased.}
      \label{sim3b}
\end{figure}

Results for Simulation \rom{3}a are given in Figure [\ref{sim3b}]. Error bars have been removed in the interest of clarity however detailed results are 
available in Table [\ref{sim3b_results}]. We note that all four algorithms struggle when $n$ is small relative to $p$. This is to be expected as 
the number of observations is much smaller than the number of parameters to be estimated. Figure [\ref{sim3b}] shows that the 
performance of the SINGLE algorithm quickly improves as $n$ increases at a rate which is similar to that of the sliding window and Gaussian kernel algorithms.

\begin{table}[ht]
  \small
  \centering
  \subfloat[DCR]{%
    \hspace{.5cm}%
\begin{tabular}{rrrrr}
  \hline
 $l$ & $\mu$ & $\sigma$   \\ 
  \hline
 10 & 0.28 & 0.09  \\ 
 20 & 0.36 & 0.15   \\ 
 30 & 0.55 & 0.21   \\ 
 40 & 0.70 & 0.15   \\ 
 50 & 0.76 & 0.08   \\ 
 60 & 0.78 & 0.07   \\ 
 70 & 0.78 & 0.06   \\ 
 80 & 0.79 & 0.03   \\ 
 90 & 0.79 & 0.02   \\ 
   \hline
\end{tabular}
    \hspace{.5cm}%
  }\hspace{1cm}
  \subfloat[SINGLE]{%
    \hspace{.5cm}%
\begin{tabular}{rrrr}
  \hline
$l$ & $\mu$ & $\sigma$   \\ 
  \hline
10 & 0.54 & 0.13   \\ 
  20 & 0.78 & 0.08   \\ 
  30 & 0.85 & 0.06 \\ 
  40 & 0.87 & 0.05  \\ 
  50 & 0.87 & 0.05  \\ 
  60 & 0.88 & 0.05  \\ 
  70 & 0.89 & 0.04   \\ 
  80 & 0.89 & 0.04   \\ 
  90 & 0.89 & 0.04   \\ 
   \hline
\end{tabular}
    \hspace{.5cm}%
  }
  
    \subfloat[Gaussian Kernel]{%
    \hspace{.5cm}%
\begin{tabular}{rrrr}
  \hline
$l$ & $\mu$ & $\sigma$   \\ 
  \hline
10 & 0.53 & 0.09  \\ 
  20 & 0.67 & 0.07    \\ 
  30 & 0.72 & 0.05    \\ 
  40 & 0.74 & 0.05    \\ 
  50 & 0.75 & 0.04    \\ 
  60 & 0.76 & 0.04    \\ 
  70 & 0.77 & 0.03    \\ 
  80 & 0.77 & 0.03    \\ 
  90 & 0.77 & 0.03    \\ 
   \hline
\end{tabular}
    \hspace{.5cm}%
  }\hspace{1cm}
    \subfloat[Sliding window]{%
    \hspace{.5cm}%
\begin{tabular}{rrrr}
  \hline
$l$ & $\mu$ & $\sigma$  \\ 
  \hline
10 & 0.41 & 0.09  \\ 
  20 & 0.49 & 0.10   \\ 
  30 & 0.55 & 0.10  \\ 
  40 & 0.61 & 0.09   \\ 
  50 & 0.65 & 0.07   \\ 
  60 & 0.67 & 0.07   \\ 
  70 & 0.68 & 0.06   \\ 
  80 & 0.69 & 0.05   \\ 
  90 & 0.69 & 0.05   \\ 
   \hline
\end{tabular}
    \hspace{.5cm}%
  }
    \caption{Detailed results from Simulation \rom{3}a. For each algorithm the mean $F$ score, $\mu$, is reported together 
    with the sample standard deviation, $\sigma$.}
    \label{sim3b_results}
\end{table}

\subsubsection{Simulation \rom{3}b - Small-world networks with decreasing $\nicefrac{n}{p}$ ratio}

As with Simulation \rom{3}a, we evaluate the performance of the proposed algorithm as the ratio of observations, $n$, relative to the 
dimensionality of the data, $p$, decreases. However, here the underlying network structure are simulated according to small-world networks. 
This simulation therefore provides an insight into how accurately proposed algorithm is able to estimate networks in the presence of rapid changes.

\begin{figure}[ht]
  \centering
    \includegraphics[width=\textwidth]{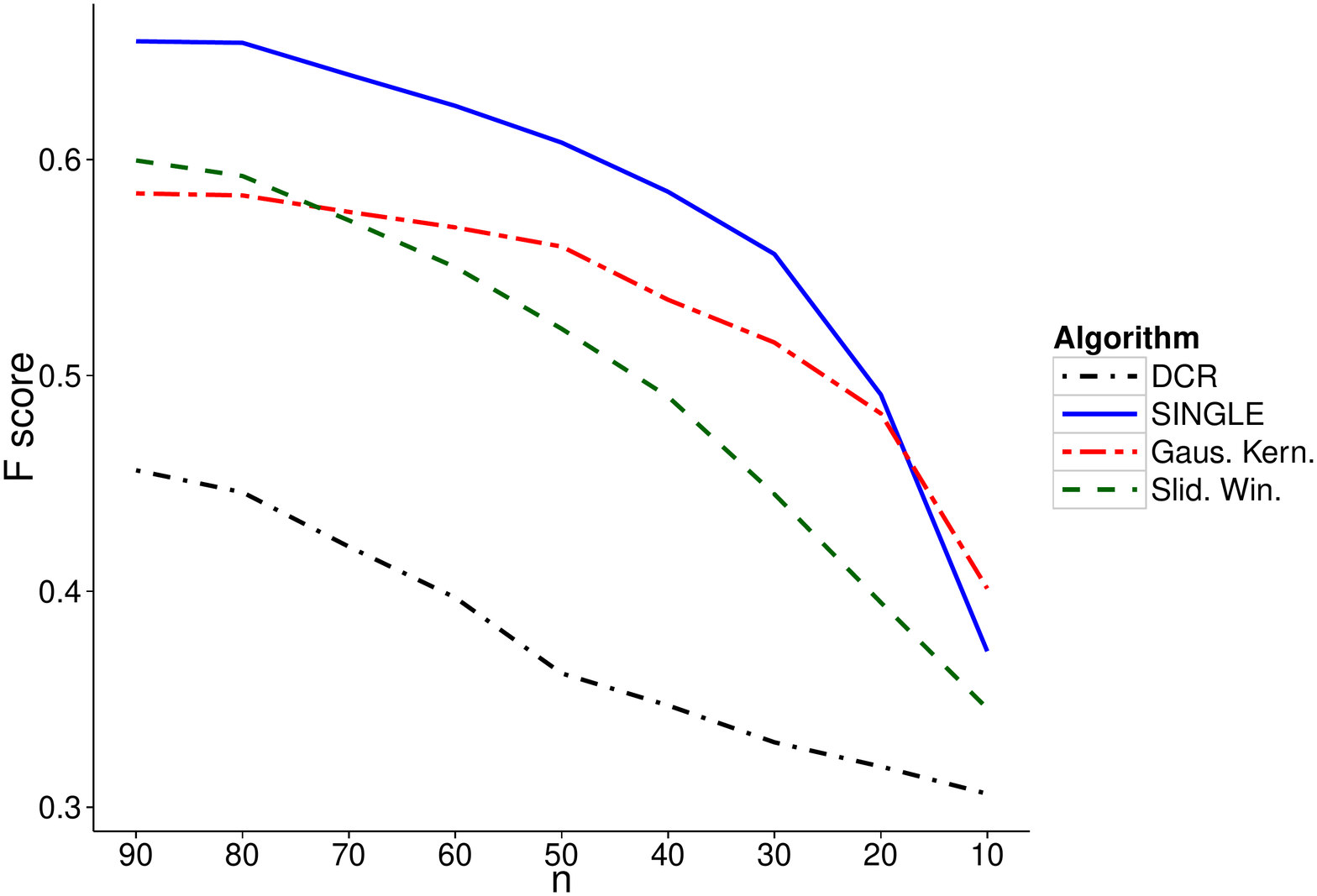}
      \caption{     
      Results for Simulation \rom{3}b for the SINGLE, DCR and sliding window algorithms respectively over 500 simulations. 
      Here networks were simulated using small-world random networks and the performance of each algorithm was studied as the ratio
      of observations, $n$, to the numbre of nodes, $p$, decreased. Here $p=10$ was fixed as $n$ decreased.
      }
      \label{sim3c}
\end{figure}

Results for Simulation \rom{3}b are given in Figure [\ref{sim3c}] and detailed results are provided in Table [\ref{sim3c_results}]. 
As with the previous simulations we note that the performance of all four algorithms is affected by the presence of small-world 
networks (see simulation \rom{1}c for a discussion). Furthermore, as in simulation \rom{3}a, the performance of all four algorithms also deteriorates 
as the ratio $\nicefrac{n}{p}$ decreases. Moreover, as in simulation \rom{3}a, the performance of the SINGLE algorithm improves as $\nicefrac{n}{p}$ increases.

\begin{table}[ht]
  \small
  \centering
  \subfloat[DCR]{%
    \hspace{.5cm}%
\begin{tabular}{rrrrr}
  \hline
 $l$ & $\mu$ & $\sigma$  \\ 
  \hline
10 & 0.31 & 0.06   \\ 
20 & 0.32 & 0.07   \\ 
30 & 0.33 & 0.08   \\ 
40 & 0.35 & 0.10   \\ 
50 & 0.36 & 0.11   \\ 
60 & 0.40 & 0.12   \\ 
70 & 0.42 & 0.11   \\ 
80 & 0.45 & 0.10   \\ 
90 & 0.46 & 0.10   \\ 
   \hline
\end{tabular}
    \hspace{.5cm}%
  }\hspace{1cm}
  \subfloat[SINGLE]{%
    \hspace{.5cm}%
\begin{tabular}{rrrr}
  \hline
$l$ & $\mu$ & $\sigma$   \\ 
  \hline
10 & 0.37 & 0.08   \\ 
20 & 0.49 & 0.07   \\ 
30 & 0.56 & 0.07   \\ 
40 & 0.59 & 0.07   \\ 
50 & 0.61 & 0.07   \\ 
60 & 0.62 & 0.07   \\ 
70 & 0.64 & 0.07   \\ 
80 & 0.65 & 0.06   \\ 
90 & 0.65 & 0.06   \\ 
   \hline
\end{tabular}
    \hspace{.5cm}%
  }
  
    \subfloat[Gaussian Kernel]{%
    \hspace{.5cm}%
\begin{tabular}{rrrr}
  \hline
$l$ & $\mu$ & $\sigma$  \\ 
  \hline
10 & 0.40 & 0.06   \\ 
20 & 0.48 & 0.05   \\ 
30 & 0.52 & 0.05  \\ 
40 & 0.54 & 0.05   \\ 
50 & 0.56 & 0.05   \\ 
60 & 0.57 & 0.05   \\ 
70 & 0.58 & 0.05   \\ 
80 & 0.58 & 0.05   \\ 
90 & 0.58 & 0.05   \\ 
   \hline
\end{tabular}
    \hspace{.5cm}%
  }\hspace{1cm}
    \subfloat[Sliding window]{%
    \hspace{.5cm}%
\begin{tabular}{rrrr}
  \hline
$l$ & $\mu$ & $\sigma$ \\ 
  \hline
10 & 0.35 & 0.07  \\ 
20 & 0.39 & 0.08   \\ 
30 & 0.44 & 0.08   \\ 
40 & 0.49 & 0.09   \\ 
50 & 0.52 & 0.08   \\ 
60 & 0.55 & 0.07  \\ 
70 & 0.57 & 0.07   \\ 
80 & 0.59 & 0.06   \\ 
90 & 0.60 & 0.06   \\ 
   \hline
\end{tabular}
    \hspace{.5cm}%
  }
    \caption{Detailed results from Simulation \rom{3}b. For each algorithm the mean $F$ score, $\mu$, is reported together 
    with the sample standard deviation, $\sigma$.}
    \label{sim3c_results}
\end{table}

\subsubsection{Computational Cost}

From a practical perspective we are also interested in the computational cost of the SINGLE algorithm. While this has already been discussed previously 
we look to benchmark the computational cost of the SINGLE algorithm relative to the previously considered algorithms.

As noted in section \ref{sec::comp_complex}, the limiting factor in the computational cost of the SINGLE algorithm is the number 
of nodes, $p$. We note that this is also the case
for the sliding window, Gaussian kernel and DCR algorithms (see Appendix \ref{app_DCR}).
As a result we compare the running times of each of the algorithms as $p$ increases for fixed $n=100$. 

Here each data set was simulated in the same manner as in Simulation \rom{1}c. That is, each dataset consisted of 3 segments of length 100 (resulting
in an overall duration of 300). The correlation structure within each segment was then randomly generated according to small-world network.

\begin{figure}[ht]
  \centering
    \includegraphics[width=\textwidth]{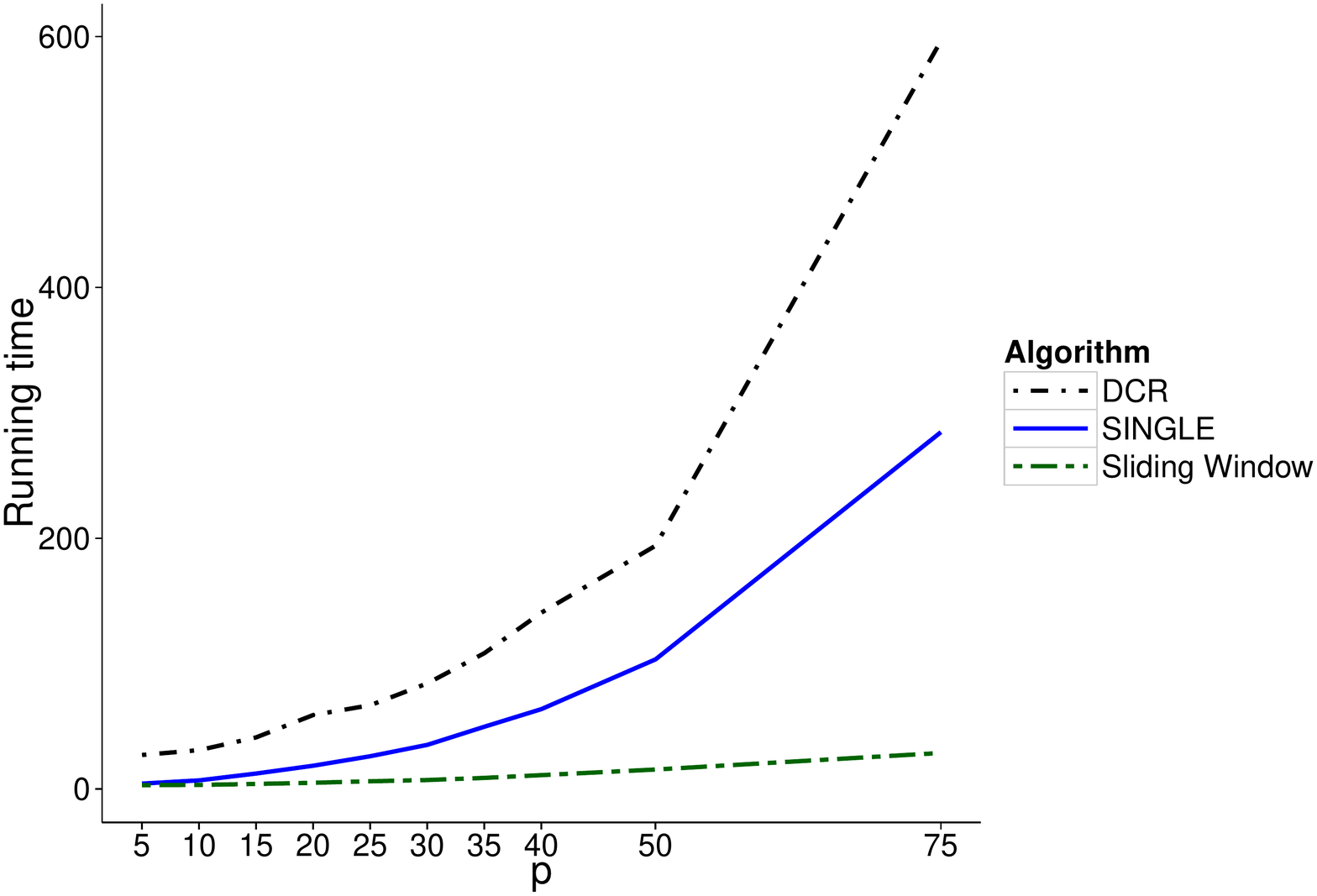}
      \caption{Average running time (seconds) for each algorithm for increasing $p$. For clarity the sliding window and Gaussian kernel
      approaches have been plotted together as they have the same computational complexity.
      }
      \label{Run_Time}
\end{figure}

In Figure [\ref{Run_Time}] we plot the mean running time of each algorithm over 50 iterations for increasing $p$. It is clear that the computational
cost of the SINGLE algorithm increases exponentially with $p$. However we note that for $p=75$ nodes the algorithm can still be run in under 5 minutes,
making it practically feasible. This simulation was run on a computer with an \textsc{Intel Core i5 CPU} at 2.8 GHz.

%

\section{Application to a Choice Reaction Time (CRT) task fMRI dataset}
\label{crt_app}

In this section we assess the ability of the SINGLE algorithm when detecting changes in real fMRI data evoked using
a simple cognitive task, the Choice Reaction Time (CRT) task. The CRT is a forced choice visuo-motor decision task that 
reliably activates visual, motor and many cognitive control regions. The task was blocked into alternating task and
rest periods. As a result we expect the task onset to evoke an abrupt change in the correlation structure that is cyclical in nature.

This is a highly challenging data set for several reasons. Firstly, it corresponds to the scenario where $\nicefrac{n}{p} = \nicefrac{126}{18}$ is 
small. Secondly, there is a high rate of change in the correlation structure with a change in cognitive state roughly every 15 seconds. 
Finally, given the nature of the CRT task 
there is a recurring correlation structure with subjects alternating between two cognitive states: resting and performing the CRT task. 
As we have seen in the simulations 
the SINGLE algorithm is well equipped to handle the aforementioned challenges.

In order to study the roles of the various ROIs during the CRT task we consider the changes in betweenness centrality of each node over 
time. The betweenness centrality of a node is the sum of how many shortest paths between 
all other nodes pass through it \citep{pandit2013traumatic}. Nodes with high betweenness centralities are considered to be of important, hub nodes in the
network \citep{hagmann2008}.

\begin{figure}[ht]
  \centering
    \includegraphics[width=\textwidth]{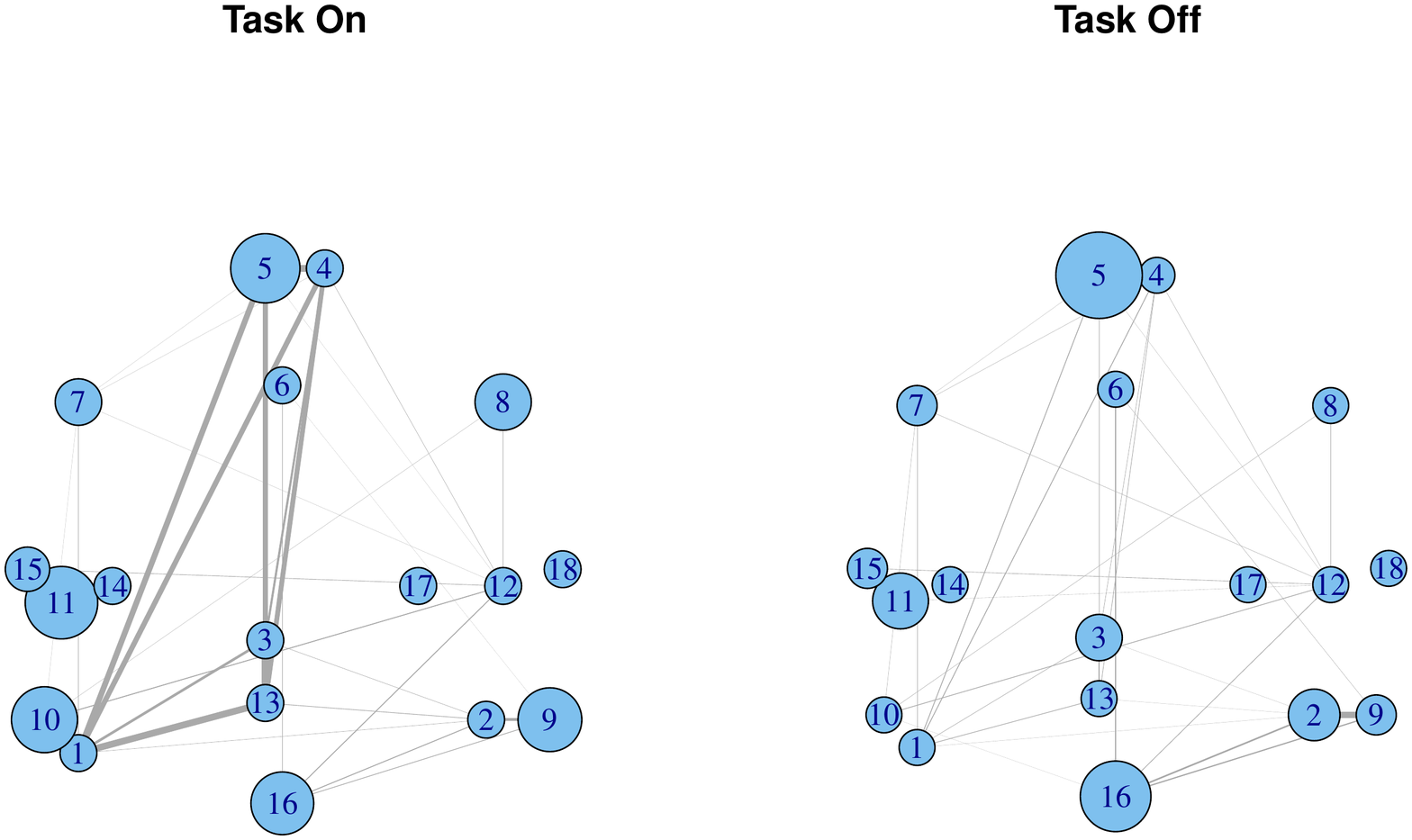}
      \caption{Mean estimated graphs on and off task for a given subject. Here node size is proportional to betweenness centrality 
      and edge width is proportional to the magnitude of their partial correlations. Each node corresponds
to a ROI given in Table [\ref{table:MNI}]. A movie of the estimated networks showing the complete evolution is available in 
the supplementary material.
      }
      \label{Pat_12}
\end{figure}

As described previously the CRT task involves subjects alternating between performing a visual stimulus task (on task) and resting state (off task). 
Figure [\ref{Pat_12}] shows the average estimated functional connectivity networks for a patient on and off task respectively. Here the size of each node 
is proportional to the sum of the betweenness centralities of the corresponding ROI and the edge thickness is proportional
to the partial correlation between nodes. We notice that the on task network is appears to be more
\textit{focused} that the off task network and this can also be seen in the corresponding example video provided in the supplementary material.

\begin{figure}[ht]
  \centering
    \includegraphics[width=\textwidth]{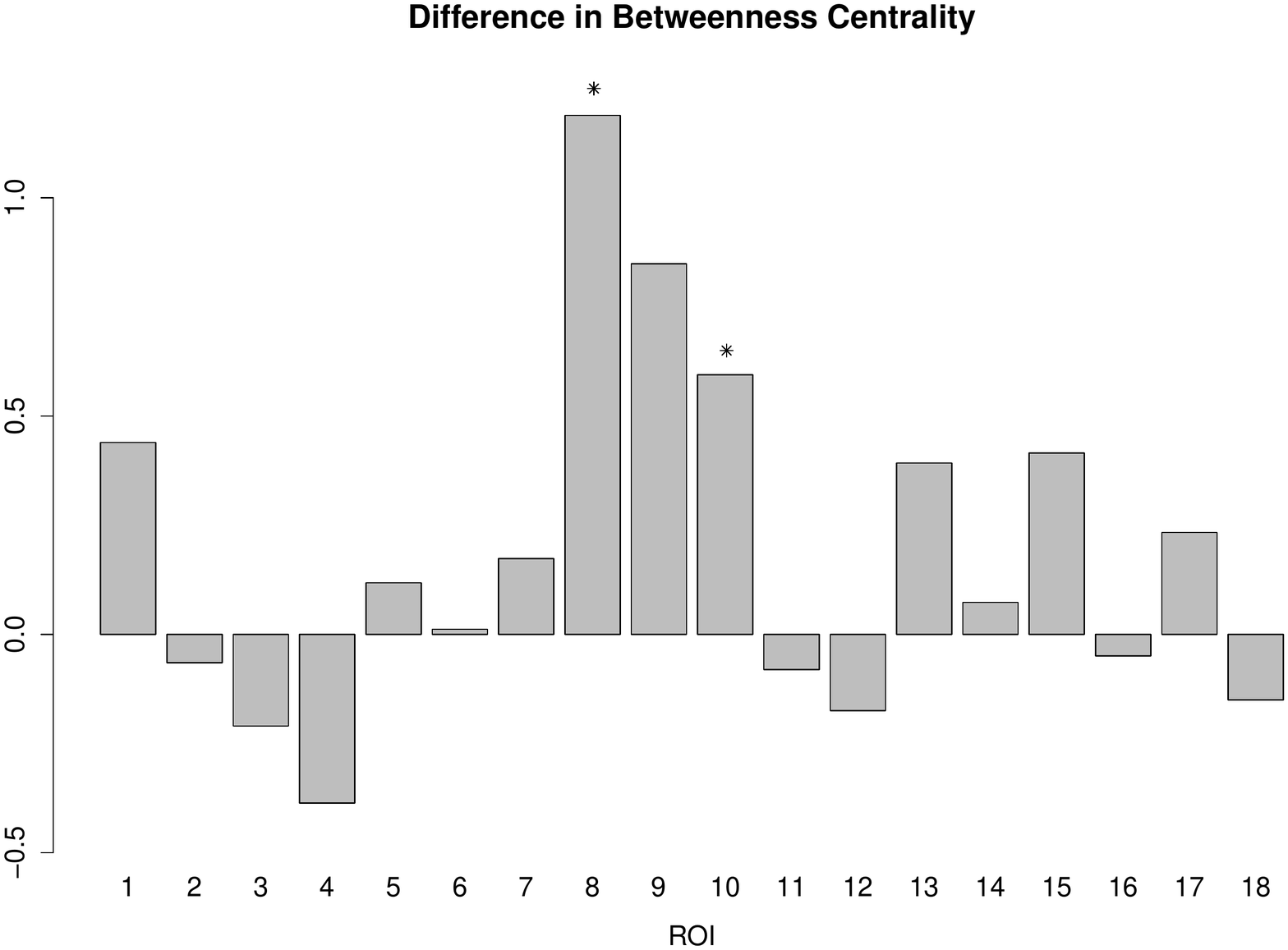}
      \caption{Estimated percentage change in betweenness centrality from off task to on task over all 24 patients. Each column corresponds to a ROI given
      in table [\ref{table:MNI}]. The * indicates a statistically significant difference in betweenness centrality at $\alpha=5\%$ level after
      correcting for multiple hypotheses.
      }
      \label{between_pic}
\end{figure}

We note that there are changes in the betweenness centralities of several nodes between tasks. In order to determine the significance of any 
changes betweenness centrality 
as a result of the changing cognitive state of the subjects 
we study the estimated graphs for each of the 24 subjects both on and off task. Figure [\ref{between_pic}] shows the percentage 
change in betweenness centrality from off task to on task for each ROI. To determine the statistical significance of reported 
changes a Wilcoxon rank sum test was employed. The resulting $p$-values where adjusted according to the  
 Bonferroni-Holm method in order to account for multiple tests. The results indicated that at 
the $\alpha=5\%$ level there was a statistically significant increase in betweenness centrality for the
8th (Right Inferior Frontal Gyrus) and 10th (Right Inferior Parietal) ROIs. This indicates that during this simple, cognitive task the 
Right Inferior Frontal Gyrus and the Right Inferior Parietal become more hub-like. This is particularly true in the case of the 
Right Inferior Frontal Gyrus where the change in betweenness centrality is particularly sizeable. 

These findings suggest that the Right Inferior Frontal Gyrus and Right Inferior Parietal play a key role in cognitive control and executive 
functions as demonstrated by their 
dynamically changing betweenness centrality throughout the task. This result agrees with the proposed functional roles for the Right Inferior
Frontal Gyrus (and adjacent right anterior insula), which is assumed to play a fundamental role in attention and executive function during cognitively 
demanding tasks and may have an important role in regulating the balance between other brain regions
\citep{Aron2003, hampshire2010, bonnelle2012}. 
The findings also agree with the proposed function of the Right Inferior Parietal lobe, which has been reported to 
play a role in high-level cognition \citep{mattingley1998motor} and sustaining attention
\citep{corbetta2002,husain2007space}.

One possible interpretation of the  the increase in betweenness centrality is that
the Right Inferior Frontal Gyrus becomes more important for the flow of information around the brain during the more challenging cognitive task.

\section{Discussion}

\label{discussion}

In this work we introduce the Smooth Incremental Graphical Lasso Estimation (SINGLE) algorithm, a new methodology 
for estimating sparse dynamic functional connectivity networks from non-stationary fMRI data.
Our approach provides two main advantages. First, the proposed algorithm is able to accurately estimate functional 
connectivity networks at each observation. This allows for the quantification the dynamic behaviour
of brain networks at a high temporal granularity. The second advantage lies in the SINGLE algorithm's 
ability to quantify network variability over time. In SINGLE, networks are estimated simultaneously in a unified 
framework which encourages temporal homogeneity. This results in networks with sparse innovations in edge structure over time and 
implies that changes in connectivity structure are only reported when substantiated by evidence in the data. 
Although the use of the SINGLE algorithm is particularly suitable for task related experiments, there is also a 
growing body of evidence to suggest 
that 
functional connectivity is highly non-stationary even in resting state, 
making the SINGLE algorithm these studies as well.

The SINGLE algorithm is closely related to sliding window based algorithms. 
We note that \cite{wasserman} have extensively studied the combined use of kernel methods and constrained optimisation to estimate
dynamic networks and provide a theoretical guarantee that accurate estimates of time varying network structure can be obtained in such 
a manner under mild assumptions. The approach taken there is to estimate sample covariance matrices at each $i \in \{1,\ldots, T\}$ using kernel methods with 
the Graphical Lasso being used subsequently to estimate the corresponding precision matrices. However, given $T$ time points this approach 
corresponds directly to $T$ independent iterations of the Graphical Lasso. As a result, while smooth estimates of the sample covariance matrix 
are obtained via the use of kernels, there is no mechanism in place to enforce temporal homogeneity in the corresponding precision matrices. Consequently
the estimated partial correlations may not accurately represent the functional connectivity over time. The SINGLE 
algorithm addresses precisely this problem by directly enforcing temporal homogeneity. This is achieved via the 
introduction an additional constraint inspired by the Fused Lasso. As shown in our simulation study, this additional constraint results 
in higher accuracy of estimated networks in a vast array of scenarios.


The SINGLE algorithm requires the input of 3 parameters, $\lambda_1, \lambda_2$ and $h$, each of which has a natural interpretation for
the user. Penalty parameters $\lambda_1$ and $\lambda_2$ enforce sparsity and temporal homogeneity respectively. They can be tuned by 
minimising $AIC$ over a given range of values. 
The choice of $h$ can be interpreted as the window length and we provide an data-driven method for tuning parameter $h$ using
the leave-one-out log-likelihood. We note that the choice of $h$ is a delicate matter as well as an active area of research in
its own right. The choice of $h$ can be seen as a trade-off between stability and temporal adaptivity. Setting $h$ to be too large will result in network
estimates that resemble the global mean and omit valuable short-term fluctuations in connectivity structure.
Conversely, setting $h$ to be too small will lead to networks that are
dominated by noise. Given this reasoning, it is often desirable to have a kernel width which is dependent on the location within the time series. This allows 
the kernel width to decrease in the proximity of a change-point (allowing for rapid temporal adaptivity) and increase when data is piece-wise stationary (in
order to fully exploit all relevant data). The idea of adaptive values of $h$ has been studied in literature such as 
\cite{AF} and \cite{principe2011kernel}, however, we leave this for future work.


Our simulation results indicate that the SINGLE algorithm can accurately estimate the true underlying
functional connectivity structure when provided with non-stationary multi-variate time series data.
We identify three relevant scenarios where the proposed method performs competitively. The first, demonstrated by simulation \rom{1},
quantifies our claim that the SINGLE algorithm is able to accurately estimate dynamic 
functional connectivity networks. 
In task based experiments it is often the case that tasks are repetitively performed followed by a period of
rest, resulting in the presence of a cyclic functional connectivity structure. This scenario is studied in simulation \rom{2} which
serves as an indication that the SINGLE algorithm is not adversely affected in such cases.
Furthermore, we have shown that the SINGLE algorithm is relatively robust
when the ratio of observations to nodes falls, meaning that the SINGLE algorithm can be applied on a subject-by-subject
basis. This is a great advantage as it avoids the issue of subject-to-subject variability and allows for the estimation of functional 
connectivity networks for each subject. This potentially allows for estimated dynamic connectivity to be used to differentiate between subjects. 
Finally, the computational cost of the proposed algorithm is studied empirically in simulation \rom{4}. A summary of all the simulation
results is provided in Table [\ref{comparison_table}].

\begin{table}[ht]
\centering
  \begin{tabular}{ |l | c | c | c|}
   \hline {} & SINGLE & DCR & Sliding window/Gaussian kernel \\ \hline \hline 
   Temporal adaptivity & $\checkmark$ & $\checkmark$ &$\checkmark$ \\ \hline
   Temporal homogeneity & $\checkmark$ & $\checkmark$  & $\text{\sffamily X}$ \\ \hline
   Cyclic correlation structure &  $\checkmark$ & $\text{\sffamily X}$ &  $\checkmark$ \\ \hline
   Parameters &$h, \lambda_1, \lambda_2$ &$\Delta, \lambda_1$& $h, \lambda_1$\\ \hline
   Computational Complexity &$\mathcal{O}(np^3 + p^2n \mbox{log}(n))$ & $\mathcal{O}((n+b)p^3)$& $\mathcal{O}(np^3)$\\ \hline 
   
  \end{tabular}
  \caption{Comparative summary of each algorithm. A derivation of the computational cost of the DCR algorithm is provided in Appendix \ref{app_DCR}
  where $b$ refers to the number of bootstrap permutation tests performed at each iteration.}
  \label{comparison_table}
\end{table}

We do note that the performance of the SINGLE algorithm was affected by the presence of 
small-world network structure. We believe this may be caused by the high local clustering present in such networks. This results in the 
short minimum path lengths between many nodes. This would cause there be a large number of correlated nodes which are not directly connected. 
It has been reported that the Lasso (and by extension the Graphical Lasso) cannot guarantee consistent variable selection in the 
presence of highly correlated predictors \citep{zou2005regularization, zou2006adaptive}. 
This issue has recently been studied in the context of genetic networks by \cite{peng2009partial} and in future these approaches could be 
adapted to address such issues.

We have presented an application showing that the SINGLE algorithm can detect cyclical changes in network structure with fMRI data acquired
while subjects perform a simple cognitive task and identify the Right Inferior Frontal Gyrus as well as the Right Inferior Parietal 
as changing their graph theoretic properties as 
the functional connectivity network reorganises. We find that there is a significant increase in the betweenness centrality for both these regions.
These findings suggest that the Right Inferior Frontal Gyrus together with the Right Inferior Parietal
play a key role in cognitive control and the functional reorganisation of brain networks. In the case of the Right Inferior Frontal Gyrus, this result
agrees with the proposed functional roles of the Right Inferior Frontal Gyrus \citep{Aron2003, corbetta2002, hampshire2010, bonnelle2012}. The Right Inferior
Parietal lobe has also been reported to play a role in high-level cognition \citep{mattingley1998motor} and sustaining attention
\citep{corbetta2002,husain2007space}.
One possible 
interpretation is that both the Right Inferior Frontal Gyrus and the Right Inferior Parietal 
become more important to the flow of information during the more challenging cognitive 
task as demonstrated by their rise in betweenness centrality.

In conclusion, the SINGLE algorithm provides an alternative and novel method for estimating the underlying network structure 
associated with dynamic fMRI data. It is ideally suited to analysing data where a change in the correlation structure is expected 
but little more is known. Going forward, the SINGLE algorithm can be applied to different types of fMRI data sets, exploring different
cognitive tasks such as those with multiple task demands, exploring how networks change with more subtle differences in cognitive state
(i.e., rather than just task on or off). Similarly, the approach can be used to investigate spontaneous network reorganisation in the 
resting state and compare this across different subject groups (e.g., comparing pathological states with healthy controls). From a methodological
point of view it would be interesting to consider variations of the SINGLE objective function, particularly with respect to the Fused Lasso 
component of the penalty. For example, this component could be exchanged with a trend filtering penalty \citep{kim2009ell_1}.

\newpage
\appendix

\section*{Appendix}
\renewcommand{\thesubsection}{\Alph{subsection}}
\label{app2}

Here we formally derive some of the results discussed in the main text. Throughout this section we assume the following results relating to convex functions:
\begin{enumerate}[ {(}1{)} ]
  \item A function $f: \mathbb{R}^{\textbf{n}} \rightarrow \mathbb{R}$ is convex if and only if the function $g:\mathbb{R}\rightarrow\mathbb{R}$ where
  $$
  g(t) = f(x + tv)
  $$
  $$ 
  \textbf{dom } g = \{ t : x+tv \in \textbf{dom } f\} 
  $$
  is convex in $t$ for all $x \in \textbf{dom } f$ and $v \in \mathbb{R}^\textbf{n}$. \\
  Here we write $\textbf{dom } f$ to denote the domain of function $f$. \label{imp_assump}
  \item Assuming $f$ is twice differentiable (i.e., its Hessian $\nabla f(x)$ exists for all $x \in \textbf{dom }f$) then $f$ will be convex if and only if its Hessian is positive semidefinite. \label{second_order_assump}
  \item The composition of convex functions is itself a convex function \label{comp_assump}
  \item Any norm is convex (this follows from the definition of a norm and the triangle inequality) \label{norm_assumption}
  \item The sum of convex functions is convex \label{add_assumption}
\end{enumerate}

\subsection{The SINGLE objective function given in equation (\ref{SINGLE_cost}) is convex}
\label{app_1}
Recall the SINGLE cost function was defined as:
\begin{equation}
 f(\{ \Theta_i \}) =  \sum_{i=1}^T -\mbox{log det } \Theta_i + \mbox{trace } ( S_i \Theta_i) + \lambda_1 \sum_{i=1}^T ||\Theta_i||_1 + \lambda_2 \sum_{|i-j|<k} ||\Theta_i - \Theta_j||_1,
\end{equation}

From Assumption (\ref{add_assumption}) it suffices to show that each component of $ f(\{ \hat \Theta \})$ is convex.
Furthermore from Assumptions  (\ref{comp_assump}) and (\ref{norm_assumption}) it follows that $||\Theta_i||_1$ and 
$||\Theta_i - \Theta_j||_1$ are convex for all $i$ and $j$. We note that 
$\mbox{trace } (\hat S_i \Theta_i) = \sum_{r=1}^p \sum_{q=1}^p (S_i)_{r,q} \cdot (\Theta_i)_{r,q}$. Therefore $\mbox{trace } (\hat S_i \Theta_i)$ 
is an affine function for all $i$ as it is a linear sum.  Finally we come to $-\mbox{log det } \Theta_i$. It follows that showing that 
$-\mbox{log det } \Theta_i$ is convex is equivalent to showing that $\mbox{log det } \Theta_i$ is concave. In order to do so we use Assumption
(\ref{imp_assump}). Formally we define $f: \mathbb{R}^{p\times p}_{++} \rightarrow \mathbb{R}$ as $f(X) = \mbox{log det } (X)$, 
where $\mathbb{R}^{p\times p}_{++}$ refers to the set of positive semi-definite $p$ by $p$ matrices. We also define $g(t) = \mbox{log det } (X + tV)$ 
for all $V \in \mathbb{R}^{p\times p}_{++}$. Since $X$ is positive semi-definite it 
follows that $X$ has a square root $X^{-\frac{1}{2}}$. Thus in order to show that $g$ is concave we can rewrite $X + tV$ as follows:
$$
X+tV  = X^{\frac{1}{2}} (I + t X^{-\frac{1}{2}} V X^{-\frac{1}{2}}) X^{\frac{1}{2}}
$$
Thus we have that $g(t) = \mbox{log det } (X + tV) = \mbox{log det } (X) + \mbox{log det } (I + t X^{-\frac{1}{2}} V X^{-\frac{1}{2}})$.

Now we can take the eigendecomposition of $t X^{-\frac{1}{2}} V X^{-\frac{1}{2}} = \Omega \Lambda \Omega^T$ where $\Omega$ is an 
orthonormal matrix of eigenvectors and $\Lambda$ is a diagonal matrix where the $i$th entry along the diagonal is the $i$th eigenvalue, 
$\lambda_i$. Finally we note that:
\begin{align*}
\mbox{log det } (I + t X^{-\frac{1}{2}} V X^{-\frac{1}{2}}) &=  \mbox{log det } (\Omega (I + t \Lambda)\Omega^T)\\
&= \sum_{i=1}^p \mbox{log } (1 + t \lambda_i)
\end{align*}
By differentiating $\mbox{log } (1 + t \lambda)$ and using Assumption (\ref{second_order_assump}) we note that $g$ is concave and thus conclude that $f(X) = \mbox{log det } (X)$ is concave and that the SINGLE cost function is convex.

\subsection{The scaled augmented Lagrangian corresponding to equations (\ref{sep1}) and (\ref{sep2}) is given by $\mathcal{L}_{\gamma} (\{\Theta_i\}, \{Z_i\}, \{U_i\})$ as shown in equation (\ref{aug_lagrange1}).}
\label{app_3}
In the case of equations (\ref{sep1}-\ref{sep2}) the corresponding Lagrangian is given by:
\small
\begin{equation}
\begin{split}
\label{laplace}
\mathcal{L} \left ( \{\Theta_i\}, \{Z_i\}, \{Y_i\} \right ) &= -\sum_{i=1}^{T} \left ( \mbox{log det } \Theta_i - \mbox{ trace } ( S_i \Theta_i) \right) \\
&+ \lambda_1 \sum_{i=1}^{T}||Z_i||_1 +\lambda_2 \sum_{i=2}^T ||Z_i - Z_{i-1}||_1 + \sum_{i=1}^T \mathrm{vec}(Y_i)^T \mathrm{vec}(\Theta_i - Z_i)
\end{split}
\end{equation}
\normalsize
where $Y_1,\ldots,Y_T, Y_i \in \mathbb{R}^{p \times p}$ are Lagrange multipliers or dual variables. The final term in the Lagrangian is equivalent to the sum of all elements in the matrix $Y_i \cdot (\Theta_i - Z_i) $. 

The augmented Lagrangian is essentially composed of the original Lagrangian and an additional penalty term. In our case the augmented Lagrangian is given by:
\small
\begin{equation}
\begin{split}
\label{aug_lagrange}
\mathcal{L} \left ( \{\Theta_i\}, \{Z_i\}, \{Y_i\} \right ) &= -\sum_{i=1}^{T} \left ( \mbox{log det } \Theta_i - \mbox{ trace } ( S_i \Theta_i) \right) + \lambda_1 \sum_{i=1}^{T}||Z_i||_1 \\
&+\lambda_2 \sum_{i=2}^T ||Z_i - Z_{i-1}||_1 + \sum_{i=1}^T \mathrm{vec}(Y_i)^T \mathrm{vec}(\Theta_i - Z_i) + \nicefrac{\gamma}{2} \sum_{i=1}^T ||\Theta_i - Z_i||_2^2
\end{split}
\end{equation}
\normalsize

We can simplify equation (\ref{aug_lagrange}) by noting that $ \mathrm{vec}(Y_i)^T \mathrm{vec}(\Theta_i - Z_i)$ is equivalent to the elementwise sum of entries of the matrix $Y_i \cdot (\Theta_i - Z_i)$ where $\cdot$ denotes the elementwise multiplication of matrices. Thus we can combine the linear and quadratic constraint terms as follows:
\begin{align}
 Y_i \cdot (\Theta_i - Z_i) + \nicefrac{\gamma}{2} || \Theta_i - Z_i ||^2_2 &= \nicefrac{\gamma}{2} || \Theta_i - Z_i + (\nicefrac{1}{\gamma}) Y_i ||_2^2 - (\nicefrac{1}{2\gamma}) ||Y_i||_2^2\\
 &= \nicefrac{\gamma}{2} || \Theta_i - Z_i + U_i ||_2^2 - \nicefrac{\gamma}{2}|| U_i||_2^2
\end{align}
where $U_i = \nicefrac{1}{\gamma}Y_i$ are the \textit{scaled} Lagrange multipliers. This yields the scaled augmented Lagrangian given in equation (\ref{aug_lagrange}). \\

\subsection{If symmetric matrices $X, Y \in \mathbb{R}^{p\times p}$ satisfy $X^{-1} - \alpha X = Y$ for some constant $\alpha$ then it follows that $X$ and $Y$ have the same eigenvectors. Furthermore it is also the case that the $i$th eigenvectors of $X$ and $Y$, denoted by $\lambda_{X_i}$ and $\lambda_{Y_i}$ respectively, will satisfy $\lambda_{X_i}^{-1} - \alpha \lambda_{X_i}  = \lambda_{Y_i}$ for $i \in \{1,\ldots, p\}$}
\label{app_2}

In order to prove claim 2 we begin taking the eigendecompositions of $X$ and $Y$ as $ \Omega_X \Lambda_X \Omega^T_X$ and $ \Omega_Y \Lambda_Y \Omega^T_Y$ respectively. Substituting these into $X^{-1} - \alpha X = Y$ we obtain:
$$ ( \Omega_X \Lambda_X \Omega^T_X)^{-1} - \alpha ( \Omega_X \Lambda_X \Omega^T_X) = \Omega_Y \Lambda_Y \Omega^T_Y $$
Expanding the left hand side yields:
\begin{align*}
\Omega_X \Lambda_X^{-1} \Omega^T_X - \alpha ( \Omega_X \Lambda_X \Omega^T_X) &= \Omega_Y \Lambda_Y \Omega^T_Y \\
&= \Omega_X (\Lambda_X^{-1} - \alpha \Lambda_X) \Omega^T_X
\end{align*}
where we have made us of the fact that $\Omega_X$ is an orthonormal matrix. Thus it follows that $\Omega_X = \Omega_Y$ and since both $\Lambda_X$ and $\Lambda_Y$ are diagonal matrices we also have that $\lambda_{X_i}^{-1} - \alpha \lambda_{X_i}  = \lambda_{Y_i}$ for $i \in \{1,\ldots, p\}$\\

\subsection{Each of the $\nicefrac{p^2}{2}$ optimisations of the form given in equation (\ref{fused_objective_thing}) can be solved by applying the Fused Lasso Signal Approximator.}
\label{app_5}
The Lasso is a regularised regression method that selects a sparse subset of predictors in least squares estimation. That is, the Lasso minimises the following objective function:
$$ \frac{1}{2} || y - X \beta ||^2 + \lambda_1 \sum_{i=1}^p |\beta_i|$$

where $y \in \mathbb{R}^{n \times 1}$ is the response vector, $X \in \mathbb{R}^{n \times p}$ is the matrix of predictors and $\beta \in \mathbb{R}^{p \times 1}$ is a vector of coefficients. The Fused Lasso extends the Lasso under the assumption that there is a natural ordering to the coefficients $\beta$. The Fused Lasso is able to do so by adding an additional penalty to the Lasso objective function as follows:
$$ \frac{1}{2} || y - X \beta ||^2 + \lambda_1 \sum_{i=1}^p |\beta_i| + \lambda_2 \sum_{i=2}^n | \beta_i - \beta_{i-1}|$$
Here only adjacent coefficients $\beta_i$ and $\beta_{i-1}$ are penalised but the Fused Lasso objective function can be specified to as to induce sparsity between any subset of $\beta$. A special case of the Fused Lasso occurs when $X = I_p$. In this case the $\lambda_2$ penalty results in $\beta$ being a piece-wise continuous approximation to $y$. We note that equation (\ref{fused_objective_thing}) resembles the objective function of the Fused Lasso. This can be seen by setting $y_i = (\Theta^j_i)_{kl} + (U^{j-1}_i)_{kl}$ and $\beta_i = (Z_i)_{kl} $ for $i=1,\ldots, T$.

\subsection{The dual update in Step 3 guarantees dual feasibility in the $\{Z_i\}$ variables and dual feasibility in the $\{\Theta_i\}$ variables can be checked by considering $||Z^{k+1}-Z^k||^2_2$.}
\label{app_4}
This result is taken from \cite{ADMM}.
Consider the general (unscaled) augmented Lagrangian with arbitrary matrices $A$, $B$ and $c$: 

$$ \mathcal{L}_{\gamma}(\Theta, Z, Y) = f(\Theta) + g(Z) + Y^T(A  \Theta + B  Z - c) + \nicefrac{\gamma}{2}||A  \Theta + B  Z - c||_2^2$$

All solutions must satisfy the following constraints:
\begin{align*}
\mbox{Primal feasibility:   } &\vspace{10mm} A  \Theta + B Z - c = 0\\
\mbox{Dual feasibility:   } &\vspace{5mm} \nabla_\Theta f(\Theta) + A^T Y = 0 \mbox{ and } \nabla_Z g(Z) + B^T Y = 0
\end{align*}
where dual feasibility is based on the unscaled, unaugmented Lagrangian.  

The ADMM algorithm iteratively minimises $\Theta$ and $Z$ such that at iteration $k+1$ we obtain $Z^{k+1}$ that minimises 
$\mathcal{L}_{\gamma}(\Theta^{k+1}, Z, Y^k)$. From this it follows that:
\small
\begin{align*}
0 &= \nabla_Z \mathcal{L}_{\gamma}(\Theta^{k+1}, Z, Y^k)\\
&= \nabla_Z \left \{f(\Theta^{k+1}) + g(Z^{k+1}) + Y^{k}(A  \Theta^{k+1} + B  Z^{k+1} - c) + \nicefrac{\gamma}{2}||A  \Theta^{k+1} + B  Z^{k+1} - c||_2^2 \right \}\\
&= \nabla_Z g(Z^{k+1}) + B^T Y^k + \gamma B^T \left(A  \Theta^{k+1} + B  Z^{k+1} -c \right )\\
&= \nabla_Z g(Z^{k+1}) + B^T \left( Y^k + \gamma \left(A  \Theta^{k+1} + B  Z^{k+1} -c \right ) \right )
\end{align*}
\normalsize

Thus it follows that by setting $Y^{k+1} = Y^k + \gamma \left(A  \Theta^{k+1} + B  Z^{k+1} -c \right )$ dual feasibility in the $Z$ variable is guaranteed. Finally, after rescaling by $U = \nicefrac{1}{\gamma}Y$ and noting that in the SINGLE algorithm $A=I_n$ and $B=-I_n$ we get the update in step 3.

Now we can continue to consider criteria for confirming dual feasibility in terms of $\{\Theta\}$ variables. Since we are guaranteed dual feasibility in $\{Z\}$ variables we only need to check for dual feasibility in $\{\Theta\}$ variables. Since $\Theta^{k+1}$ minimises $\mathcal{L}_{\gamma}(\Theta, Z^k, Y^k)$ we have:

\begin{align*}
 0 &=  \nabla_{\Theta} \mathcal{L}_{\gamma}(\Theta, Z^k, Y^k)\\
   &=  \nabla_{\Theta} f(\Theta^{k+1}) + A^T \left (Y^k + \gamma \left (A  \Theta^{k+1} + B  Z^{k} -c \right ) \right )\\
   &=  \nabla_{\Theta} f(\Theta^{k+1}) + A^T \left (\underbrace{Y^k + \gamma (A  \Theta^{k+1} + + B Z^{k+1} -c)}_{Y^{k+1}} + \gamma (B  Z^{k} - B Z^{k+1})\right ) \\
   &= \nabla_{\Theta} f(\Theta^{k+1}) + A^T Y^{k+1} + \gamma A^T B (Z^k - Z^{k+1})
\end{align*}

Thus in order to have dual feasibility in $\{\Theta\}$ variables we require 
$$ \gamma A^T B (Z^{k+1} - Z^{k}) = \nabla_{\Theta} f(\Theta^{k+1}) + A^T Y^{k+1}.$$ 
Since in our case we have that $A = I_n$ and $B=-I_n$ it follows that we can check for dual feasibility by considering $||Z^{k+1}-Z^k||_2^2$.

\subsection{The computational complexity of the DCR algorithm is $\mathcal{O}((n+b)p^3)$ where $b$ is the number of bootstrap permutations to perform}
\label{app_DCR}

We begin by noting that the computational complexity of the Graphical Lasso is $\mathcal{O}(p^3)$. While it is possible to 
reduce the computational complexity in some special cases we do not consider this below.

Prior to outlining our proof, a brief overview of the DCR algorithm is in order. The DCR algorithm looks to estimate 
dynamic functional connectivity networks by segmenting data into piece-wise continuous partitions. Within each partition the 
network structure is assumed to be stationary, allowing for the use of a wide variety of network estimation algorithms. In the case of the DCR 
the Graphical Lasso is chosen. 

The data, $\{ X_i \in \mathbb{R}^{1 \times p}: i=1, \ldots, T\}$, is segmented using a greedy partitioning scheme. Here the global network is first
estimated using the Graphical Lasso. The BIC of the global network is noted and subsequently used to propose change-points. The DCR algorithm then 
proceeds to partition the data into subsets $A_{\gamma} = \{ X_i: i=1, \ldots, \gamma\}$ and $B_{\gamma} = \{ X_i: i=\gamma +1, \ldots, T\}$ for 
$\gamma \in \{\Delta+1\, \ldots, T- \Delta\}$. Thus $\Delta$ represents the minimum number of observations between change-points. 

For each of these partitions a network is estimated for $A_\gamma$ and $B_\gamma$ and their joint BIC is noted. This step therefore involves $\mathcal{O}(n)$ 
iterations of the Graphical Lasso, resulting in a computation complexity of $\mathcal{O}(np^3)$.

Subsequently, the value of $\gamma$ resulting in the greatest reduction in BIC relative to the global network is proposed as a change-point. In order to check
the statistical significance of the proposed change-point a block bootstrap permutation permutation test is performed. This step involves a further $b$ 
iterations of the Graphical Lasso where $b$ is the number of bootstrap permutations performed. As a result this step has a computational complexity
of $\mathcal{O}(bp^3)$.

This procedure is repeated until all significant change-points have been reported. We therefore conclude that the computational complexity of the 
DCR algorithm is $\mathcal{O}((n+b)p^3)$.

\newpage
\bibliographystyle{plainnat}
\bibliography{ref}

\end{document}